\documentclass[letterpaper, 10 pt, journal, twoside]{ieeeconf}  % Comment this line out if you need a4paper
\usepackage{cite}
\usepackage{amsmath,amssymb,amsfonts}
\allowdisplaybreaks[4]
\usepackage[caption=false,font=normalsize,labelfont=sf,textfont=sf]{subfig}
\usepackage{algorithmic}
\usepackage{graphicx}
\usepackage{textcomp}
\usepackage{xcolor}
\usepackage{bm}

\IEEEoverridecommandlockouts                              % This command is only needed if 
\title{\LARGE \bf
Generating 6-D  Trajectories for Omnidirectional Multirotor Aerial Vehicles in Cluttered Environments
}

\author{Peiyan Liu, Yuanzhe Shen, Yueqian Liu, Fengyu Quan, Can Wang, and Haoyao Chen% <-this % stops a space
\thanks{This work was supported in part by the National Natural Science Foundation of China under Grant U1713206 and Grant 61673131. (Corresponding author: Haoyao Chen. E-mail: hychen5@hit.edu.cn.)}% <-this % stops a space
\thanks{P.Y. Liu, Y.Z. Shen, F.Y. Quan, and H.Y. Chen* are with the School of Mechanical Engineering and Automation, Harbin Institute of Technology Shenzhen, P.R. China. Y.Q. Liu is with the Faculty of Aerospace Engineering, Delft University of Technology, Netherland. C. Wang is with Shenzhen Institute of Advanced Technology, Chinese Academy of Sciences, Shenzhen, P.R. China. }%
}

\begin{document}

\maketitle
%\thispagestyle{empty}
%\pagestyle{empty}

%%%%%%%%%%%%%%%%%%%%%%%%%%%%%%%%%%%%%%%%%%%%%%%%%%%%%%%%%%%%%%%%%%%%%%%%%%%%%%%%
\begin{abstract}
    As fully-actuated systems, omnidirectional multirotor aerial vehicles (OMAVs) have more flexible maneuverability and advantages in aggressive flight in cluttered environments than traditional underactuated MAVs. %Due to the high dimensionality of configuration space, making the designed trajectory generation algorithm efficient is challenging.
    This paper aims to achieve safe flight of OMAVs in cluttered environments. 
    Considering existing static obstacles, an efficient optimization-based framework is proposed to generate 6-D  $SE(3)$ trajectories for OMAVs. 
    Given the kinodynamic constraints and the 3D collision-free region represented by a series of intersecting convex polyhedra, the proposed method finally generates a safe and dynamically feasible 6-D  trajectory. 
    First, we parameterize the vehicle's attitude into a free 3D vector using stereographic projection to eliminate the constraints inherent in the $SO(3)$ manifold,  
    while the complete $SE(3)$ trajectory is represented as a 6-D polynomial in time without inherent constraints. 
    The vehicle's shape is modeled as a cuboid attached to the body frame to achieve whole-body collision evaluation. 
    Then, we formulate the origin trajectory generation problem as a constrained optimization problem. 
    The original constrained problem is finally transformed into an unconstrained one that can be solved efficiently. 
    To verify the proposed framework's performance, simulations and real-world experiments based on a tilt-rotor hexarotor aerial vehicle are carried out.
\end{abstract}

\section{Introduction}

%\subsection{Background}
Multirotor aerial vehicles (MAVs) have stood out from various intelligent robots and entered our lives from laboratories. 
However, most traditional MAVs are underactuated systems, which means their translation and rotation dynamics are coupled.
This nature limits the maneuverability of traditional MAVs.
In order to fully exploit the potential of MAVs, 
several kinds of omnidirectional MAVs (OMAVs) with decoupled position and attitude control have been developed in recent years.
By changing the configuration of rotors \cite{brescianini2016design} or adding tilting degrees of freedom to rotors \cite{kamel2018voliro}, 
this kind of MAV can perform controlled and free rigid body motion, which is impossible for traditional underactuated ones.
In some extreme scenarios, such as a narrow straight passage, 
traditional MAVs coupling acceleration with attitude will be most likely unable to pass through it without collision,
while OMAVs can tilt themselves to adapt to the narrow space by controlling the attitude and simultaneously, control its position to achieve smooth and collision-free passing. % (Fig. \ref{fig:narrow_passage_demo}.
Such advantages make OMAVs bound to play a great application value in scenarios like aerial manipulation and disaster rescue.

\begin{figure}[t]
    \centering
    \includegraphics[width=3.4in]{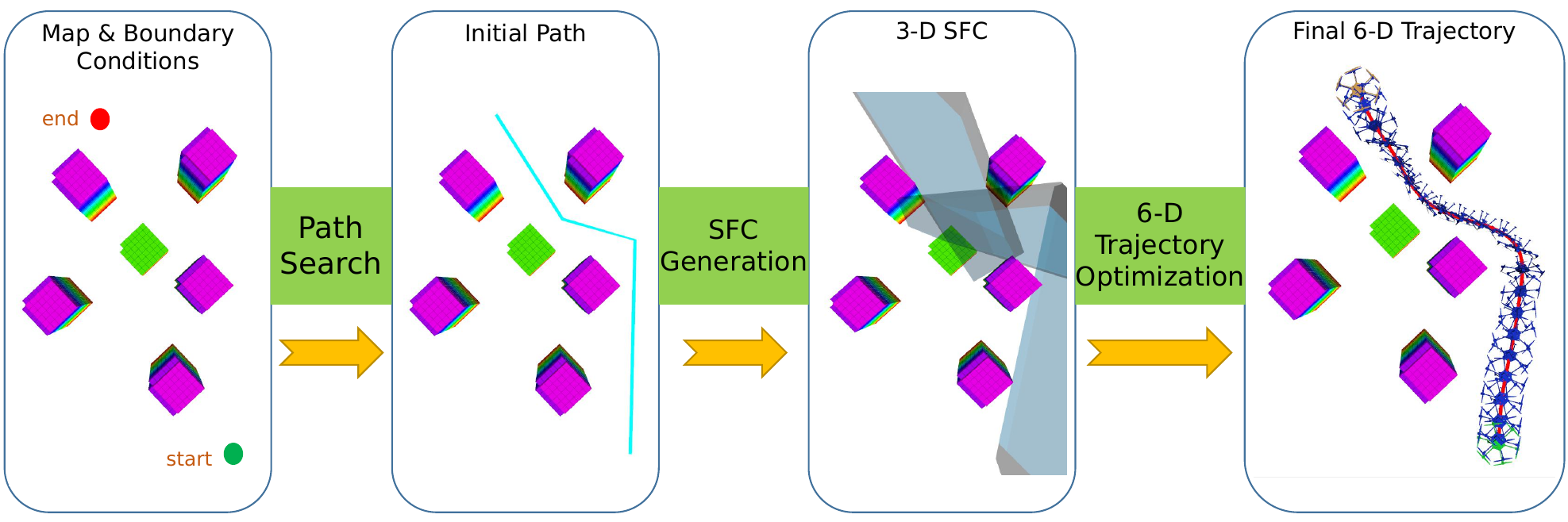}
    \caption{The proposed 3-stage 6-D trajectory generation framework involves initial path search, SFC generation, and 6-D trajectory optimization. }
\label{fig:framework}
\end{figure}

%\subsection{Motivations}
In order to better exploit the potential of OMAVs, a proper trajectory generation framework is indispensable.
It needs to generate 6-D  $SE(3)$ trajectories and take into account the vehicle's shape and pose to adapt to cluttered environment.
Moreover, the lightweight requirements of MAVs limit the available onboard computing power. So we also expect the designed framework to have lower computational burdens and higher computational efficiency than traditional methods. 
%To achieve this goal, we are faced with the following challenges: %(1) Due to the high dimensionality of the configuration space $SE(3)$ (6 degrees of freedom), generating trajectories using either search-based or optimization-based methods is prone to high computational time;
%(1) The configuration space $SE(3)$ is a non-Euclidean manifold in which it is difficult to describe the collision-free region; 
%(2) Moreover, finding an appropriate way to represent the rotation trajectory is necessary to make the trajectory generation problem easy to solve.

% \begin{itemize}
%     \item Due to the high dimensionality of the configuration space $SE(3)$ (6 degrees of freedom), generating trajectories using search-based or optimization-based methods is prone to high computational time.
%     \item The configuration space $SE(3)$ is a non-Euclidean manifold in which it is difficult to describe the collision-free region. Moreover, it is necessary to find an appropriate way to represent the attitude trajectory to make the trajectory generation problem easy to solve.
% \end{itemize}

% \begin{figure}[!t]
%     \centering
%     \includegraphics[width=3.3in]{figures/narrow_passage_situation.png}
%     \caption{The visualization of generated 6-D  collision-free trajectory through a narrow passage for OMAVs using our method.}
% \label{fig:narrow_passage_situation}
% \end{figure}

%\subsection{Related Work}
The existing research on OMAVs mainly focuses on the mechanical structure design and control strategy, but few achievements have been made in trajectory generation.
Brescianini \emph{et al.} efficiently generate 6-D trajectories that satisfy certain input constraints for OMAVs using motion primitives.
An energy-efficient trajectory generation method for a tilt-rotor hexarotor UAV is proposed in \cite{morbidi2018energy}. 
Pantic \emph{et al.} \cite{pantic2021mesh} present a trajectory generation method based on Riemannian Motion Policies (RMPs), which aims to drive a vehicle to fly to and along a specified surface and is applied to aerial physical interaction. 
The above works do not take into account the obstacles in the environment.
%Nguyen \emph{et al.} \cite{nguyen2016time} present a 6-D  $SE(3)$ collision-free trajectory generation method for spacecraft in which RRT \cite{lavalle1998rapidly} is applied to search for a collision-free initial path in $SE(3)$ state space and then time-optimal path parameterization (TOPP) is used to obtain a collision-free 6-D  trajectory.
%The method in \cite{nguyen2016time} only directly parameterizes the segmented path searched by RRT, and the start and end velocities of each segment need to be 0, resulting in poor trajectory smoothness, and the iterative shortcutting step is quite inefficient.

The existing works on trajectory generation of OMAVs do not meet our requirements well, while trajectory generation methods in the position space $\mathbb{R}^3$ are relatively mature for traditional underactuated MAVs. 
Mellinger \textit{et al.} \cite{mellinger2011minimum} generate smooth trajectories by minimizing the square integral of the trajectory derivatives for the first time.   
Several efficient schemes have been created based on the idea in \cite{mellinger2011minimum}. 
In order to meet the safety requirements in a cluttered environment, gradient information in the map is used in \cite{gao2017gradient} and \cite{zhou2020ego} to push the trajectories away from obstacles to achieve collision avoidance. 
Another common way is to use intersecting geometry primitives to approximate the free space connecting the start and goal points  \cite{gao2020teach}\cite{gao2019flying}\cite{yang2021whole}\cite{han2021fast}, 
and the union of these primitives is called a safe flight corridor (SFC).
Wang \textit{et al.} \cite{wang2022geometrically} propose an optimization-based trajectory generation framework for multicopters.
It shows state-of-the-art performance in efficiency, extensibility, and solution quality.
Based on \cite{wang2022geometrically},  several whole-body $SE(3)$ trajectory generation methods are proposed \cite{yang2021whole}\cite{han2021fast}\cite{ren2023online} in which whole-body safety constraints can be constructed conveniently and handled efficiently with 3-D free space approximated by the polyhedral SFC. 
However, these methods are only suitable for under-actuated MAVs whose position and attitude are highly coupled. 

%\subsection{Contributions}
This paper proposes an efficient 3-stage 6-D trajectory generation framework for OMAVs in cluttered environments, as shown in Fig. \ref{fig:framework}. 
Similar to \cite{han2021fast}, 3-D polyhedral SFC is used to represent the collision-free region.
First, we use sampling-based methods like RRT \cite{lavalle1998rapidly} to search for an initial feasible path from the start to the end point and generate a 3-D SFC based on the initial path. 
Then, the 6-D trajectory optimization stage efficiently generates a smooth, safe, and dynamically feasible 6-D trajectory connecting the start and the end states with linear time and space complexity. 
For trajectory representation, we represent a vehicle's attitude as a Hamilton unit quaternion $\mathbf{Q} \in \mathbb{S}^3$ \cite{sola2017quaternion} and parameterize it into a free 3-D vector using stereographic projection to eliminate the constraints inherent in the $\mathbb{S}^3$ manifold. 
This 3-D attitude vector is combined with the 3-D position vector to form a 6-D pose vector.
The complete $SE(3)$ trajectory is then represented as a piece-wise polynomial of the pose vector over time. 
The vehicle's shape and attitude are taken into account to apply whole-body safety constraints to the trajectory.
We formulate the trajectory generation problem as a constrained optimization problem and finally transform it into an unconstrained one that can be solved using quasi-Newton methods. 

In summary, the main contributions of this paper are: 
\begin{itemize}
    % \item We adopt a stereographic-projection-based rotation parameterization to represent rotation trajectories as curves in $\mathbb{R}^3$. 
    \item Based on 3-D rotation parameterization and the idea of the state-of-the-art trajectory optimization method \cite{wang2022geometrically} for traditional underactuated MAVs, we represent rotation trajectories as curves in $\mathbb{R}^3$ and present a 6-D rigid body trajectory optimization framework considering whole-body safety and dynamic limits for OMAVs.  
    \item We achieve efficient trajectory optimization by adopting a stereographic-projection-based rotation parameterization. 
    \item We propose an efficient 6-D trajectory generation pipeline that leverages the full potential of OMAVs for obstacle avoidance for the first time. 
\end{itemize}

\begin{figure}[t]
    \centering
    \includegraphics[width=3.1in]{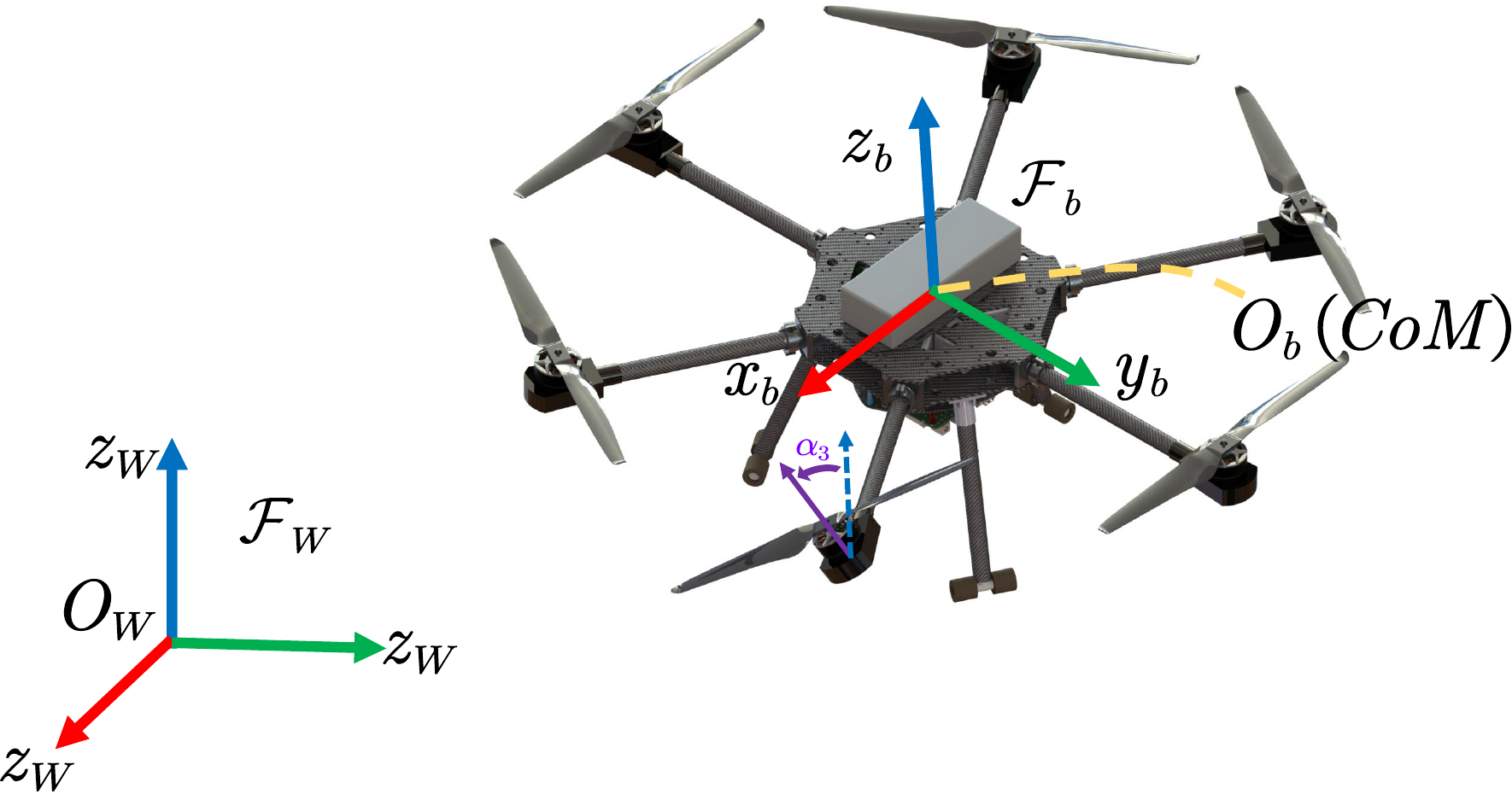}
    \caption{This figure shows the frame definitions, the rotor tilt angle, and the CAD model of OmniHex. $\mathcal{F}_W$ uses the east-north-up (ENU) coordinate system. 
    The origin of $\mathcal{F}_b$ coincides with the vehicle's Center of Mass (CoM);  
    the $x_b$ axis points forward, the $y_b$ axis points to the body's left, and the right-hand rule determines the $z_b$ axis. }
\label{fig:frame_definitions}
\end{figure}

\section{System modeling and control}

\subsection{Definitions}

The methodology of this paper mainly involves two right-handed coordinate systems: the world (inertial) frame $\mathcal{F}_W$ and the body frame $\mathcal{F}_b$ (Fig. \ref{fig:frame_definitions}). 
We denote $\mathbf{a}_A$ the coordinate of a vector $\mathbf{a}$ expressed in frame $\mathcal{F}_A$, and we omit the subscript if $\mathcal{F}_A$ is the world frame $\mathcal{F}_W$.
Denote $\mathbf{R}$ the rotation matrix of the body frame $\mathcal{F}_b$ w.r.t. the world frame $\mathcal{F}_W$,
and thus $\mathbf{a} = \mathbf{R}\mathbf{a}_b$. 

%\subsection{Differential Flatness}\label{subsec:differential_flatness}
\subsection{System Modeling of OMAVs}
\label{subsec:system_modelling}

%Differential flatness of underactuated multirotor vehicles have been well studied \cite{mellinger2011minimum}.
%We will show that OMAVs also have similar properties. 

A common OMAV has six independent control degrees of freedom,
we take joint thrust and torque generated by the rotors expressed in $\mathcal{F}_b$ as its control input 
\begin{equation}
    \mathbf{u} := 
    \begin{bmatrix}
        \mathbf{f}^{\top}_b & \bm{\tau}^{\top}_b
    \end{bmatrix}^{\top} \in \mathbb{R}^6, 
    \label{equ:control_input}
\end{equation}
and we select CoM position $\mathbf{p}$, CoM velocity $\mathbf{v} = \dot{\mathbf{p}}$, $\mathcal{F}_b$'s orientation $\mathbf{Q}$ (expressed as Hamilton unit quaternion \cite{sola2017quaternion}), and $\mathcal{F}_b$'s angular velocity $\bm{\omega}$ (all expressed in $\mathcal{F}_W$) as its state variables:
\begin{equation}
        \mathbf{x} := 
        [
            \underbrace{p_x, p_y, p_z}_{\mathbf{p}^\top}, \underbrace{Q_w, Q_x, Q_y, Q_z}_{\mathbf{Q}^\top}, \underbrace{v_x, v_y, v_z}_{\mathbf{v}^\top}, \underbrace{\omega_x, \omega_y, \omega_z}_{\bm{\omega}^\top}
        ]^{\top}. 
    \label{equ:state_variable}
\end{equation}
where $\mathbf{x} \in \mathbb{R}^{13}$ is the full state vector. 
The system's output is given by the position of its CoM and the orientation of $\mathcal{F}_b$ expressed in $\mathcal{F}_W$:
\begin{equation}
    \mathbf{y}(t) := \{\mathbf{p}(t), \mathbf{Q}(t)\} \in SE(3).
    \label{equ:trajectory_y}
\end{equation}

Expressing a rotation trajectory as a function of a unit quaternion or a rotation matrix w.r.t. time $t$ is intuitive. 
However, this implies equality constraints inherent in $SO(3)$, which can be troublesome for trajectory optimization. 
So, we consider parameterizing each rotation $\mathbf{Q}$ as an unconstrained 3-D vector. 
Denote a rotation trajectory by a curve $\bm{\sigma}(t) : [t_0, t_M] \rightarrow \mathbb{R}^3$,
and the corresponding $\mathbf{Q}(t)$ is determined by a smooth surjection $\mathbf{Q}(\bm{\sigma}) : \mathbb{R}^3 \rightarrow SO(3)$.
Then, we express the 6-D trajectory as 
\begin{equation}
    \mathbf{z}(t) := 
    \begin{bmatrix}
        \mathbf{p}^{\top}(t) & \bm{\sigma}^{\top}(t)
    \end{bmatrix}^{\top} \in \mathbb{R}^6. 
    \label{equ:trajectory_z}
\end{equation}

Now the angular velocity $\bm{\omega}$ of $\mathcal{F}_b$ can be obtained from $\bm{\sigma}$ and its finite derivatives:
\begin{equation}
    \bm{\omega}^{\wedge} = \bm{{\omega}}^{\wedge}(\bm{\sigma}, \dot{\bm{\sigma}})
                    = \sum_{i=1}^3 \frac{\partial \mathbf{R}(\bm{\sigma})}{\partial \sigma_i} \dot{\sigma}_i \mathbf{R}(\bm{\sigma})^{\top},  
    \label{equ:angular_velocity_wedge}
\end{equation}
where $\mathbf{R}$ is the rotation matrix corresponding to $\mathbf{Q}$; 
$(\cdot)^{\wedge} : \mathbb{R}^3 \rightarrow \mathfrak{so}(3)$ denotes the skew-symmetric matrix of a 3-D vector and thus $\mathbf{a} \times \mathbf{b} = \mathbf{a}^{\wedge}\mathbf{b}, \forall \mathbf{a}, \mathbf{b} \in \mathbb{R}^3$. 
The inverse map is denote as $(\cdot)^{\vee} : \mathfrak{so}(3) \rightarrow \mathbb{R}^3$. 
The angular acceleration $\dot{\bm{\omega}}$ can be further calculated according to \eqref{equ:angular_velocity_wedge} :
\begin{equation}
    \begin{aligned}
        (\dot{\bm{\omega}})^{\wedge} = \sum_{i=1}^3 \left(\sum_{j=1}^3 \frac{\partial^2 \mathbf{R}}{\partial\sigma_i\partial\sigma_j}\ddot{\sigma}_j\dot{\sigma}_i + \frac{\partial \mathbf{R}}{\partial \sigma_i}\ddot{\sigma}_i\right)\mathbf{R}^{\top} + 
        \dot{\mathbf{R}}\dot{\mathbf{R}}^{\top}. 
    \end{aligned}
    \label{equ:angular_acceleration_wedge}
\end{equation}
Thus, we obtain the expression of the state $\mathbf{x}$ w.r.t. $\mathbf{z}$ and $\mathbf{z}$'s finite derivatives, given as 
\begin{equation}
    \mathbf{x} = 
    \begin{bmatrix}
        \mathbf{p} \\ \mathbf{Q} \\ \mathbf{v} \\ \bm{\omega}
    \end{bmatrix} = 
    \Psi_x(\mathbf{z}, \dot{\mathbf{z}}) := 
    \begin{bmatrix}
        \mathbf{p} \\ \mathbf{Q}(\bm{\sigma}) \\
        \dot{\mathbf{p}} \\ 
        \left(\sum_{i=1}^3 \frac{\partial \mathbf{R}}{\partial \sigma_i} \dot{\sigma}_i \mathbf{R}^{\top}\right)^{\vee}
    \end{bmatrix}. 
    \label{equ:Psi_x}
\end{equation}
The expression of the control input $\mathbf{u}$ w.r.t. $\mathbf{z}$ and $\mathbf{z}$'s finite derivatives can be determined by combining \eqref{equ:angular_velocity_wedge} and \eqref{equ:angular_acceleration_wedge} with Newton-Euler equation, given as
\begin{equation}
    \mathbf{u} = 
    \begin{bmatrix}
        \mathbf{f}_b \\ \bm{\tau}_b
    \end{bmatrix} = 
    \Psi_u(\mathbf{z}, \dot{\mathbf{z}}, \ddot{\mathbf{z}}) := 
    \begin{bmatrix}
        m \mathbf{R}(\bm{\sigma})^{\top} (\ddot{\mathbf{p}} - \mathbf{g}) \\ 
        \mathbf{R}(\bm{\sigma})^{\top} \left(\bm{\omega}^{\wedge}\mathbf{J}(\bm{\sigma})\bm{\omega} + \mathbf{J}(\bm{\sigma})\dot{\bm{\omega}}\right)
    \end{bmatrix}, 
    \label{equ:Psi_u}
\end{equation}
where $m$ is the mass of the system;
$\mathbf{g} = \begin{bmatrix} 0 & 0 & -9.8\end{bmatrix}^\top\text{m}\cdot\text{s}^{-2}$ is acceleration of gravity in $\mathcal{F}_W$;
$\mathbf{J}(\bm{\sigma}) := \mathbf{R}(\bm{\sigma}) \mathbf{J}_b \mathbf{R}(\bm{\sigma})^{\top} \in \mathbb{R}^{3\times3}$, 
where $\mathbf{J}_b \in \mathbb{R}^{3\times3}$ is the vehicle's inertia matrix in $\mathcal{F}_b$ which can be treated as a constant.

%Differential flatness \cite{fliess1995flatness} is a fundamental property in robot motion planning, 
%which allows us to obtain the system state $\mathbf{x}$ and control input $\mathbf{u}$ from a carefully sellected system output $\mathbf{y}_f$ and its finite derivatives, and $\mathbf{y}_f$ is called a flat output.
% Planning trajectory in flat output space can not only reduce the dimension of the trajectory to simplify the problem and speed up the solution but also make the trajectory meet given constraints more easily, which is more conducive to the execution of the robot.
Equations \eqref{equ:Psi_x} and \eqref{equ:Psi_u} indicate that our selected trajectory representation $\mathbf{z}(t)$ has the properties as a flat output \cite{fliess1995flatness}. 
Therefore, it is convenient to transform constraints on the state $\mathbf{x} \in \mathbb{R}^{13}$ and the input $\mathbf{u} \in \mathbb{R}^6$ to constraints on $\mathbf{z} \in \mathbb{R}^6$ and $\mathbf{z}$'s finite derivatives using Equ. \eqref{equ:Psi_x} and Equ. \eqref{equ:Psi_u}. 

\subsection{Position and Attitude Control} \label{subsec:control}

\begin{figure}[t]
    \centering
    \includegraphics[width=3.3in]{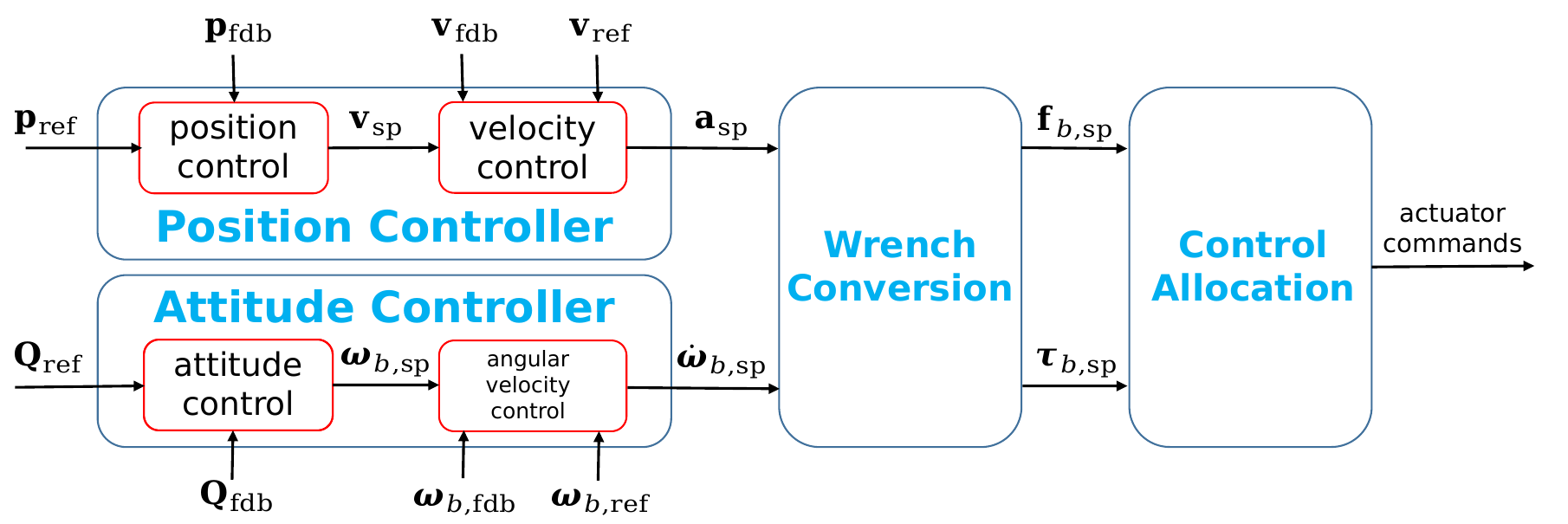}
    \caption{The control pipeline of OmniHex. As the position control and attitude control of OmniHex is decoupled, the linear and angular acceleration setpoints are calculated separately by their respective cascade PID controllers. }
\label{fig:control_pipeline}
\end{figure}

In this section, 
we briefly introduce the control strategy designed for a tilt-rotor omnidirectional hexacopter (from now on called OmniHex) to accurately track 6-D trajectories. 
As shown in Fig. \ref{fig:frame_definitions}, compared to traditional hexacopters, OmniHex adds an additional controllable degree of freedom for each rotor to rotate around the axis of the arm on which it is mounted, which allows OmniHex to generate thrust and torque in any direction relative to $\mathcal{F}_b$ for independent control of position and attitude. 
The control pipeline of OmniHex is shown in Fig. \ref{fig:control_pipeline}. 
In the following statements, the quantities with subscript ``ref'' are the setpoints given by the reference trajectory, and those with subscript ``fdb'' are the feedback from the state estimator. 

\emph{1) Position Controller: } 
The outer loop of the position controller is a proportional controller that sets the desired velocity $\mathbf{v}_{\text{sp}}$ for the inner loop based on position error $\bm{e}_{p} := \mathbf{p}_{\text{ref}} - \mathbf{p}_{\text{fdb}}$. 
\begin{equation}
\label{eq:posctrl}
    \mathbf{v}_{\text{sp}} = \mathbf{K}_{P,p} \bm{e}_{p}, 
\end{equation}
$\mathbf{K}_{P,p} \in \mathbb{R}^{3\times3}$ is the diagonal gain matrix. The inner velocity control loop uses PID control with a feed-forward design, given as 
\begin{equation}
    \label{eq:velctrl}
    \mathbf{a}_{\text{sp}} = \mathbf{K}_{P, v} \bm{e}_{v} + \mathbf{K}_{I, v} \int \bm{e}_{v}\text{d}t + \mathbf{K}_{D, v} \frac{\text{d} \bm{e}_{v}}{\text{d}t} + \mathbf{K}_{F, v} \mathbf{v}_{\text{ref}}, 
\end{equation}
where $\mathbf{K}_{P, v}, \mathbf{K}_{I, v}, \mathbf{K}_{D, v}, \mathbf{K}_{F, v} \in \mathbb{R}^{3\times3}$ are diagonal gain matrices for proportional, integral, derivative, and feed-forward control, respectively. $\bm{e}_{v} := \bm{v}_{\text{sp}} - \bm{v}_{\text{fdb}}$ is the velocity error. 

\emph{2) Attitude Control: } 
We use Hamilton unit quaternions to represent the attitudes. 
The attitude error is defined as the difference between the reference attitude $\mathbf{Q}_\text{ref}$ and the attitude feedback $\mathbf{Q}_\text{fdb}$
\begin{equation}
    \label{eq:atterr} 
    \bm{e}_{Q} := 
    \begin{bmatrix}
        e_{Q, w} & \bm{e}_{Q, \text{vec}}^\top
    \end{bmatrix}^\top = \mathbf{Q}_{\text{ref}} \otimes \mathbf{Q}_{\text{fdb}}^{-1}, 
\end{equation}
where $e_{Q, w} \in \mathbb{R}$ and $\bm{e}_{Q, \text{vec}} \in \mathbb{R}^3$ are scalar and vector parts of quaternion $\bm{e}_{Q}$, respectively. 
$\otimes$ denotes quaternion product. 
Then the outer loop of attitude control maps the attitude error to the desired angular velocity $\bm{\omega}_{b, \text{sp}}$ (expressed in $\mathcal{F}_b$) as follows: 
\begin{equation}
    \label{eq:omegasp}
    \bm{\omega}_{b, \text{sp}} = \text{sign}(e_{Q, w}) \cdot \mathbf{K}_{P, Q} \bm{e}_{Q, \text{vec}}, 
\end{equation}
Denoting angular velocity error as $\bm{e}_{\omega} := \bm{\omega}_{b, \text{ref}} - \bm{\omega}_{b, \text{fdb}}$, the inner loop calculates the desired angular acceleration as follows: 
\begin{equation}
    \begin{aligned}
    \label{eq:angacc}
    \dot{\bm{\omega}}_{b, \text{sp}} &= \mathbf{K}_{P, \omega} \bm{e}_{\omega} + \mathbf{K}_{I, \omega} \int \bm{e}_{\omega}\text{d}t + \mathbf{K}_{D, \omega} \frac{\text{d} \bm{e}_{\omega}}{\text{d}t}\\ & + \mathbf{K}_{F, \omega} \bm{\omega}_{b, \text{ref}}. 
\end{aligned}
\end{equation}

\emph{3) Wrench Conversion and Control Allocation: }
Obtaining the desired accelerations, we can calculate the desired control wrench $\mathbf{w}_{b, \text{sp}}$ (expressed in $\mathcal{F}_b$) as follows: 
\begin{equation}
    \label{eq:wrench} 
    \mathbf{w}_{b, \text{sp}} := 
    \begin{bmatrix}
        \mathbf{f}_{b, \text{sp}} \\ \bm{\tau}_{b, \text{sp}}
    \end{bmatrix} = 
    \begin{bmatrix}
        m \mathbf{Q}_{\text{fdb}}^{-1} \otimes (\mathbf{a}_{\text{sp}} - \mathbf{g}) \otimes \mathbf{Q}_{\text{fdb}} \\ 
        \mathbf{J}_b \dot{\bm{\omega}}_{b, \text{sp}} + \mathbf{d}_{\text{com}}^{\wedge}\mathbf{f}_{b, \text{sp}} + \bm{\omega}_{b, \text{fdb}}^\wedge\mathbf{J}_b\bm{\omega}_{b, \text{fdb}}
    \end{bmatrix}, 
\end{equation} 
where the offset of the CoM $\mathbf{d}_{\text{com}}$ is predefined. 
The actuator commands of OmniHex are rotation speeds $\omega_i$ and tilt angles $\alpha_i, i=1, \cdots, 6$ of its six rotors. 
Defining $\cos{\alpha_i} = c_i, \sin{\alpha_i} = s_i, i = 1, \cdots, 6$, $\mathbf{w}_{b, \text{sp}}$ calculated in \eqref{eq:wrench} can also be expressed as 
\begin{equation}
    \label{eq:alloc1}
        \mathbf{w}_{b, \text{sp}}
    =
    \bm{A}[\omega_1^2s_1,\omega_1^2c_1, \cdots, \omega_6^2s_6,\omega_6^2c_6]^\top, 
\end{equation}
where $\bm{A} \in \mathbb{R}^{6 \times 12}$ is the allocation matrix defined in \cite{bodie2020towards}. Using the strategy presented in \cite{bodie2020towards}, we can obtain the desired actuator commands.

\section{method}  \label{sec:method}
This section presents the details of the proposed 3-stage optimization-based trajectory generation framework based on the above system modeling. 
The path search and the SFC generation stages are introduced briefly in Section \ref{subsec:safety_constraint}. 
The trajectory optimization stage is described in detail in Sections \ref{subsec:problem_formulation} to \ref{subsec:rotation_parameterization}, which is the main contribution of this paper. 

\subsection{Whole-body Safety Constraint}\label{subsec:safety_constraint}

% In many existing trajectory planning schemes for multirotor vehicle like \cite{deits2015computing, gao2020teach, han2021fast}, 
To construct whole-body safety constraints, we use convex polyhedral SFC to represent the collision-free regions in 3-D position space connecting the start point $\mathbf{p}_o \in \mathbb{R}^3$ and the target point $\mathbf{p}_f \in \mathbb{R}^3$. 
First, we search an initial collision-free path $\mathbf{p}_0 \rightarrow \mathbf{p}_1 \rightarrow \cdots \rightarrow \mathbf{p}_{M_\mathcal{P}}$ from $\mathbf{p}_0 = \mathbf{p}_o$ to $\mathbf{p}_{M_\mathcal{P}} = \mathbf{p}_f$. 
Then, we use RILS \cite{liu2017planning} to find a 3-D convex polyhedron $\mathcal{P}_i$ approximating the collision-free region around each path segment $\mathbf{p}_{i-1} \rightarrow \mathbf{p}_{i}, i = 1, \cdots, M_{\mathcal{P}}$ as a primitive of SFC. 
The resulting SFC is defined as $\mathcal{SFC} := \bigcup_{i=1}^{M_\mathcal{P}}\mathcal{P}_i$. 
Two adjacent convex polyhedral primitives satisfy the connection condition:
\begin{equation}
    \left(\mathcal{P}_{i} \cap \mathcal{P}_{i + 1}\right)^\circ \neq \emptyset, i = 1, \cdots, M_{\mathcal{P}} - 1
    \label{equ:connection_condition_of_sfc}, 
\end{equation}
where $(A)^\circ$ denotes the interior of the point set $A$.

Similar to \cite{han2021fast}, the shape of the vehicle is approximated by a convex polyhedron $\mathcal{P}_\text{S}$ that is fixed to $\mathcal{F}_b$ and wraps the entire vehicle. 
The coordinates of its vertices in $\mathcal{F}_b$ denoted by $\tilde{\bm{v}}_l, l=1, \cdots L_{\text{v}}$ are known constants.
The safety of the vehicle can be guaranteed as long as the $L_{\text{v}}$ vertices are all in the convex polyhedron $\mathcal{P}$ that represents the safe region. 
We express $\mathcal{P}$ using linear inequalities:
\begin{equation}
    \mathcal{P} := \{\mathbf{p} \in \mathbb{R}^3 \vert \mathbf{n}_k^{\top}\mathbf{p} - d_k \leq 0, k = 1, \cdots, K\}, 
    \label{equ:H_representation_of_P}
\end{equation}
which means that the convex polyhedron $\mathcal{P}$ is the intersection of $K$ halfspaces.
$\mathbf{n}_k \in \mathbb{R}^3$ is the unit outer normal vector of the $k$-th halfspace. %, and $d_k$ denotes the distance from the origin of $\mathbb{R}^3$ to the boundary plane of the $k$-th half-space. 
Then the safety condition of the vehicle with position $\mathbf{p}$ and attitude $\mathbf{R}$ can be written as follow 
\begin{equation}
    \mathbf{n}_k^{\top}\bm{v}_l - d_k \leq 0, \forall k \in \{1, \cdots, K\}, \forall l \in \{1, \cdots, L_{\text{v}}\}
    \label{equ:safety_condition_at_p_R}, 
\end{equation}
where $\bm{v}_l := \mathbf{p} + \mathbf{R}\tilde{\bm{v}}_l, l=1, \cdots L_{\text{v}}$ are coordinates of vertices of $\mathcal{P}_\text{S}$ in $\mathcal{F}_W$.

%This article does not cover the details of the SFC generation method.

\subsection{6-D Trajectory Optimization} \label{subsec:problem_formulation}

For the convenience of controlling the derivatives, we express $\mathbf{z}(t)$ as a 6-D piece-wise polynomial over time: 
\begin{equation}
    \mathbf{z}(t) = \mathbf{c}_i^{\top}\bm{\beta}(t - t_{i - 1}), \text{if}\ t \in [t_{i - 1}, t_i], i = 1, \cdots, M
    \label{equ:kth_M_piece_polynomial}
\end{equation}
where $\mathbf{c}_i \in \mathbb{R}^{(n+1) \times 6}$ is the coefficient matrix of the $i$-th piece and
$\bm{\beta}(\alpha) := \begin{bmatrix}1 & \alpha & \cdots & \alpha^n\end{bmatrix}^{\top} \in \mathbb{R}^{(n+1)}$. 
$M$ is the number of pieces. 
The coefficient matrix of the whole trajectory $\mathbf{c}$ and the times allocated for all the pieces are defined as 
\begin{align}
    &\mathbf{c} := 
    \begin{bmatrix}\mathbf{c}_1^{\top} & \cdots & \mathbf{c}_M^{\top}\end{bmatrix}^{\top} \in \mathbb{R}^{M(n+1) \times 6}, \\
    & \begin{aligned}
        \mathbf{T} &:= \begin{bmatrix}t_1 - t_0 & \cdots & t_M - t_{M-1}\end{bmatrix}^{\top} \\ 
        &= \begin{bmatrix}T_1 & \cdots & T_M\end{bmatrix}^{\top} \in \mathbb{R}_+^M. 
    \end{aligned}
\end{align}
%The coefficient matrix of the whole trajectory is $\mathbf{c} := \begin{bmatrix}\mathbf{c}_1^{\top} & \cdots & \mathbf{c}_M^{\top}\end{bmatrix}^{\top} \in \mathbb{R}^{M(n+1) \times 6}$.
%The times allocated for all the pieces are $\mathbf{T} := \begin{bmatrix}t_1 - t_0 & \cdots & t_M - t_{M-1}\end{bmatrix}^{\top} = \begin{bmatrix}T_1 & \cdots & T_M\end{bmatrix}^{\top} \in \mathbb{R}_+^M$.

Our goal is to find the optimal $\mathbf{c}$ and $\mathbf{T}$ that minimize a given objective function. 
We adopt MINCO trajectory representation proposed in \cite{wang2022geometrically} to deal with the high dimensionality of $\mathbf{c}$ which may reduce optimization efficiency. 
By enforcing the optimality conditions in Theorem 2 of \cite{wang2022geometrically}, the parameters of $\mathbf{z}(t)$ can be transformed from $(\mathbf{c}, \mathbf{T})$ to the waypoints $\mathbf{q} := \begin{bmatrix}\mathbf{q}_1 & \cdots & \mathbf{q}_{M-1}\end{bmatrix} \in \mathbb{R}^{6 \times (M - 1)}, i = 1, \cdots, M-1$ and $\mathbf{T}$. 
After specifying $(\mathbf{q}$, $\mathbf{T})$, the start condition $\bar{\mathbf{z}}_o := \mathbf{z}^{[s - 1]}(t_0) := \begin{bmatrix} \mathbf{z}(t_0)^\top & \cdots & \mathbf{z}^{(s-1)}(t_0)^\top
\end{bmatrix}^\top \in \mathbb{R}^{6s}$, and the end condition $\bar{\mathbf{z}}_f  := \mathbf{z}^{[s - 1]}(t_0) := \begin{bmatrix} \mathbf{z}(t_M)^\top & \cdots & \mathbf{z}^{(s-1)}(t_M)^\top
\end{bmatrix}^\top \in \mathbb{R}^{6s}$, $\mathbf{c}$ will be determined uniquely in an efficient way by solving the sparse linear system    
\begin{equation}
    \label{equ:banded_system}
    \mathbf{M}(\mathbf{T})\mathbf{c}(\mathbf{q}, \mathbf{T}) = \mathbf{b}(\mathbf{q}) \Rightarrow \mathbf{c}(\mathbf{q}, \mathbf{T}) = \mathbf{M}^{-1}(\mathbf{T})\mathbf{b}(\mathbf{q}), 
\end{equation}
where $\mathbf{M} \in \mathbb{R}^{2Ms \times 2Ms}$ and $\mathbf{b} \in \mathbb{R}^{2Ms \times 6}$ are defined in \cite{wang2022geometrically}. 
The resulting trajectory satisfies $\mathbf{z}(t_i) = \mathbf{q}_i$ and is $2s - 2$ times continuously differentiable at $t_i, i = 1, \cdots, M - 1$. 
The degree of polynomial $\mathbf{z}(t)$ is determined by the system order $s \in \mathbb{N}_+$ as $n = 2s - 1$. 

For the convenience of the following statement, 
we divide $\mathbf{c}$ and $\mathbf{q}$ into position blocks and attitude blocks:
$ \mathbf{c}_i = 
\begin{bmatrix}
    \mathbf{c}_i^{p} & \mathbf{c}_i^{\sigma}
\end{bmatrix}, \mathbf{c}_i^{p}, \mathbf{c}_i^{\sigma} \in \mathbb{R}^{2s \times 3} $; 
$ \mathbf{c} = 
\begin{bmatrix}
    \mathbf{c}^{p} & \mathbf{c}^{\sigma}
\end{bmatrix}, \mathbf{c}^{p}, \mathbf{c}^{\sigma} \in \mathbb{R}^{2Ms \times 3} $; 
$ \mathbf{q}_i = 
\begin{bmatrix}
    {\mathbf{q}_i^{p}}^{\top} & {\mathbf{q}_i^{\sigma}}^{\top}
\end{bmatrix}^{\top}, \mathbf{q}_i^{p}, \mathbf{q}_i^{\sigma} \in \mathbb{R}^{3} $; 
$ \mathbf{q} = 
\begin{bmatrix}
    {\mathbf{q}^{p}}^{\top} & {\mathbf{q}^{\sigma}}^{\top}
\end{bmatrix}^{\top}, \mathbf{q}^{p}, \mathbf{q}^{\sigma} \in \mathbb{R}^{3 \times (M-1)} $.
% \begin{align}
%     & \mathbf{c}_i = 
%     \begin{bmatrix}
%         \mathbf{c}_i^{p} & \mathbf{c}_i^{\sigma}
%     \end{bmatrix}; \mathbf{c}_i^{p}, \mathbf{c}_i^{\sigma} \in \mathbb{R}^{2s \times 3} \\
%     & \mathbf{c} = 
%     \begin{bmatrix}
%         \mathbf{c}^{p} & \mathbf{c}^{\sigma}
%     \end{bmatrix}; \mathbf{c}^{p}, \mathbf{c}^{\sigma} \in \mathbb{R}^{2Ms \times 3} \\
%     & \mathbf{q}_i = 
%     \begin{bmatrix}
%         {\mathbf{q}_i^{p}}^{\top} & {\mathbf{q}_i^{\sigma}}^{\top}
%     \end{bmatrix}^{\top}; \mathbf{q}_i^{p}, \mathbf{q}_i^{\sigma} \in \mathbb{R}^{3} \\
%     & \mathbf{q} = 
%     \begin{bmatrix}
%         {\mathbf{q}^{p}}^{\top} & {\mathbf{q}^{\sigma}}^{\top}
%     \end{bmatrix}^{\top}; \mathbf{q}^{p}, \mathbf{q}^{\sigma} \in \mathbb{R}^{3 \times (M-1)} 
% \end{align}

\begin{figure}[t]
    \centering
    \includegraphics[width=3.3in]{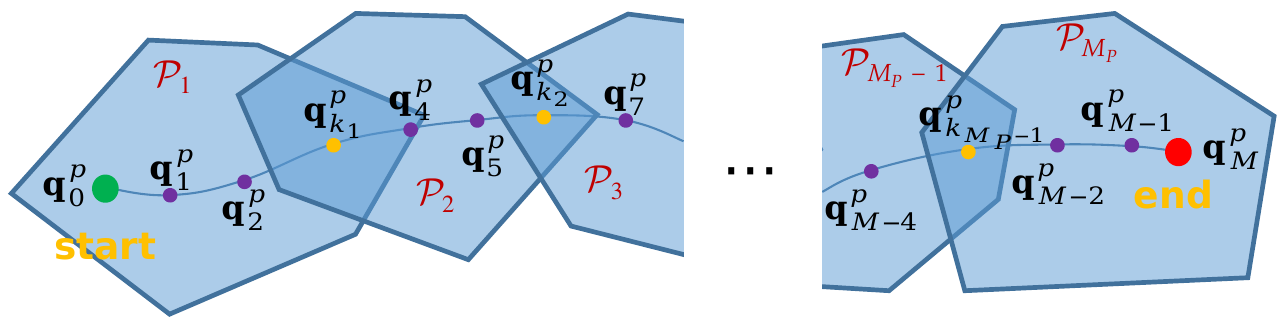}
    \caption{The 2-D illustration of spatial constraints \eqref{equ:origin_problem_cons_3}-\eqref{equ:origin_problem_cons_5}. In this case $k_1 = 3, k_2 = 6$, and $k_{M_{\mathcal{P} - 1}} = M - 3$. }
\label{fig:spatial_constraints}
\end{figure}

We expect the trajectory $\mathbf{z}(t)$ to be smooth enough and satisfy the boundary conditions, the dynamic constraints, and the safety constraints.
Then, the original form of our trajectory optimization problem is as follows:
\begin{subequations}
    \begin{align}
        \min_{\mathbf{q}, \mathbf{T}} &\int_{t_0}^{t_M} \Vert \mathbf{z}^{(s)}(t) \Vert_2^2 \text{d}t + k_\rho \Vert \mathbf{T} \Vert_1,  \label{equ:origin_problem_obj}\\
           s.t. \quad & \mathbf{z}^{[s - 1]}(t_0) = \bar{\mathbf{z}}_o, \mathbf{z}^{[s - 1]}(t_M) = \bar{\mathbf{z}}_f,  \label{equ:origin_problem_cons_2}\\
            &  \begin{aligned}
                & \mathbf{z}(t) = \mathbf{c}_i^{\top}(\mathbf{q}, \mathbf{T})\bm{\beta}(t - t_{i - 1}), \forall t \in [t_{i - 1}, t_i], \\
                & i = 1, \cdots, M,  \label{equ:origin_problem_cons_1}
            \end{aligned} \\
             &\begin{aligned}
                & \mathbf{q}_{k_i}^p \in \left(\mathcal{P}_{i} \cap \mathcal{P}_{i + 1}\right)^\circ, i = 1, \cdots, M_{\mathcal{P}} - 1,  \label{equ:origin_problem_cons_3}
             \end{aligned} \\
             & \begin{aligned}
                & \mathbf{q}_j^p \in \mathcal{P}_1, \text{if} \ 1 \leq j < k_1,   \\
                & \mathbf{q}_j^p \in \mathcal{P}_{M_\mathcal{P}}, \text{if} \ k_{M_{\mathcal{P}} - 1} < j \leq M - 1,  \label{equ:origin_problem_cons_4}
             \end{aligned} \\
             & \mathbf{q}_j^p \in \mathcal{P}_i, \text{if} \ k_{i-1} < j < k_i, i = 2, \cdots, M_{\mathcal{P}} - 1, \label{equ:origin_problem_cons_5}\\
             & \mathbf{T} \succ \mathbf{0},  \label{equ:origin_problem_cons_6}\\
             & \Vert \dot{\mathbf{p}}(t) \Vert_2^2 \leq v_{\text{max}}^2, \forall t \in [t_0, t_M],  \label{equ:origin_problem_cons_7}\\
             & \Vert \ddot{\mathbf{p}}(t) \Vert_2^2 \leq a_{\text{max}}^2, \forall t \in [t_0, t_M],  \label{equ:origin_problem_cons_8}\\
             & \Vert \bm{\omega}(t) \Vert_2^2 \leq \omega_{\text{max}}^2, \forall t \in [t_0, t_M],  \label{equ:origin_problem_cons_9}\\
             & \bm{v}_l(t) \in \mathcal{P}^i, \forall t \in [t_{i - 1}, t_i], i = 1, \cdots M; l = 1, \cdots, L_{\text{v}}. \label{equ:origin_problem_cons_10}
    \end{align}
    \label{equ:origin_problem}
\end{subequations}
The first term of the objective function \eqref{equ:origin_problem_obj} is the smoothness cost, from now on denoted as $J$, and the second is the time regularization term. 
Spatial constraints \eqref{equ:origin_problem_cons_3}-\eqref{equ:origin_problem_cons_5} bind the position waypoints to a specific region in the SFC, where $k_i, i = 1, \cdots, M_{\mathcal{P}}$ is the indices of waypoints that should be confined to $\left(\mathcal{P}_{i} \cap \mathcal{P}_{i + 1}\right)^\circ$ and $ 1 \leq k_i < k_{i + 1} \leq M - 1,  i = 1, \cdots, M_{\mathcal{P}} - 1$, as illustrated in Fig. \ref{fig:spatial_constraints}. %This can prevent the trajectory from deviating from the SFC during the optimization. 
Temporal constraint \eqref{equ:origin_problem_cons_6} ensures that the time allocated to each piece is strictly positive. 
Inequalities \eqref{equ:origin_problem_cons_7}-\eqref{equ:origin_problem_cons_9} are kinodynamic constraints according to the task requirements and vehicle limits; 
Safety constraint \eqref{equ:origin_problem_cons_10} confines each trajectory piece to a polyhedron primitive of SFC.
We let $\mathcal{P}^i := \{\mathbf{p} \in \mathbb{R}^3 \vert \mathbf{n}_{i, k}^{\top}\mathbf{p} - d_{i, k} \leq 0, k = 1, \cdots, K_i\}$ denote the polyhedron to which the $i$-th piece is assigned. 
Note that in addition to \eqref{equ:origin_problem_cons_7}-\eqref{equ:origin_problem_cons_10}, we can also add more continuous time constraints according to actual needs, such as dealing with the vehicle's actuator delays or dynamic changes in the environment.
%\begin{equation}
%    \mathcal{P}^i = \{\mathbf{p} \in \mathbb{R}^3 \vert \mathbf{n}_{i, k}^{\top}\mathbf{p} - d_{i, k} \leq 0, \Vert \mathbf{n}_{i, k} \Vert_2 = 1, k = 1, \cdots, K_i\}
%\end{equation}

%For the selection of order $s$, according to \eqref{equ:Psi_u},
%$\mathbf{z}(t)$ needs to be at least three-order continuously differentiable to ensure the smoothness of the control input. 
%That is, $n = 2s - 1 \geq 4$, which gives $s \geq 3$.

The original trajectory optimization problem \eqref{equ:origin_problem} contains various constraints. 
To deal with them, we draw on the ideas in \cite{wang2022geometrically}. 
The continuous-time constraints \eqref{equ:origin_problem_cons_7} to \eqref{equ:origin_problem_cons_10} can be softened as integral penalty terms. 
The spatial constraints \eqref{equ:origin_problem_cons_3}-\eqref{equ:origin_problem_cons_5} and temporal constraint \eqref{equ:origin_problem_cons_6} can be eliminated using certain diffeomorphisms $\mathbf{q}^p(\bm{\xi})$ and $\mathbf{T} (\bm{\tau})$ suggested in \cite{wang2022geometrically}. 
Note that different from \cite{wang2022geometrically}, our optimization variables include rotation waypoints $\mathbf{q}^{\sigma}$. 
Our strategy is not to impose any hard constraints on $\mathbf{q}^{\sigma}$, so there is no need to transform it. 
Finally, what we need to solve is an unconstrained optimization problem as follows:
    \begin{align}
        \min_{\bm{\xi}, \mathbf{q}^{\sigma}, \bm{\tau}} 
        &J(\mathbf{q}^p(\bm{\xi}), \mathbf{q}^{\sigma}, \mathbf{T}(\bm{\tau})) + k_\rho \Vert \mathbf{T}(\bm{\tau}) \Vert_1 \nonumber \\
        &+ \mathcal{W}_v \sum_{i=1}^M \sum_{j=1}^{\kappa} \mathcal{V}\left(\left\| \dot{\mathbf{p}}\left(\hat{t}_{ij}\right) \right\|_2^2 - v_{\text{max}}^2\right) \frac{T_i}{\kappa} \nonumber \\
        &+ \mathcal{W}_a\sum_{i=1}^M \sum_{j=1}^{\kappa} \mathcal{V}\left(\left\| \ddot{\mathbf{p}}\left(\hat{t}_{ij}\right) \right\|_2^2 - a_{\text{max}}^2\right) \frac{T_i}{\kappa} \nonumber \\
        &+ \mathcal{W}_{\omega} \sum_{i=1}^M \sum_{j=1}^{\kappa} \mathcal{V}\left(\left\| \bm{\omega}\left(\hat{t}_{ij}\right) \right\|_2^2 - \omega_{\text{max}}^2\right) \frac{T_i}{\kappa} \nonumber \\
        &+ \mathcal{W}_c \sum_{i=1}^M \sum_{j=1}^{\kappa} \sum_{l=1}^{L_{\text{v}}} \sum_{k=1}^{K_i} \mathcal{V}\left(\mathbf{n}_{i, k}^{\top}\bm{v}_l\left(\hat{t}_{ij}\right) - d_{i, k}\right) \frac{T_i}{\kappa},  
        \label{equ:final_uncons_opt_problem}
    \end{align}
where $\mathcal{W}_\star$ is the weight of the corresponding penalty term. 
$\kappa \in \mathbb{N}_+$ controls the resolution of numerical integration. 
$\mathcal{V}(\cdot):=\max(\cdot, 0)^3$ measures the constraint violation on the trajectory at the sampling time $\hat{t}_{ij} := t_{i-1} + \frac{j}{\kappa}T_i$. 
With closed-form gradients, problem \eqref{equ:final_uncons_opt_problem} can be efficiently solved by quasi-Newton methods. 
%It should be noted that the continuous time constraints softened and discretized above may be slightly violated, 
%but in most cases, these violations are within the acceptable range.

% \subsection{Gradient Calculation} \label{subsec:general_gradient_calculation}
% The gradients of the objective function w.r.t. the optimization variables are needed to solve the unconstrained optimization problem \eqref{equ:final_uncons_opt_problem}. 
% First, we calculate the gradients w.r.t. $\mathbf{c}$ and $\mathbf{T}$ (refer to Appendix \ref{app:gradient_calculation}), 
% Then, the gradients w.r.t. optimization variables $\bm{\xi}$, $\mathbf{q}^\sigma$, and $\bm{\tau}$ can be obtained using the formulas derived as \cite{wang2022geometrically}.

% Since the rotation-related quantities such as $\mathbf{R}$ and $\bm{\omega}$ are closely related to the rotation vector $\bm{\sigma}$ defined in Section \ref{subsec:system_modelling}, 
% the evaluation of these penalty terms and their gradients will vary depending on the rotation parameterization map $\mathbf{R}(\bm{\sigma}): \mathbb{R}^3 \rightarrow SO(3)$.

\subsection{Rotation Parameterization} \label{subsec:rotation_parameterization}
From Section \ref{subsec:problem_formulation} we can find that choosing an appropriate rotation parameterization map $\mathbf{R}(\bm{\sigma})$ is essential for our trajectory generation.
There are several commonly used ways to parameterize a rotation in $SO(3)$ as a vector in $\mathbb{R}^3$,
such as Euler angles and axis-angle (also known as a Lie algebra of $SO(3)$). 
However, since $\bm{\sigma}$ is a free vector on which we do not apply any hard constraints, 
it will be ambiguous if $\bm{\sigma}$ represents Euler angles or an axis-angle: 
two very different $\bm{\sigma}$ values will most likely correspond to the same rotation.
Moreover, Euler angle representation has the problem of gimbal lock. 

Considering these shortcomings, 
Euler angle or an axis-angle representation lacks rationality when using polynomial interpolation.
In our implementation, 
we adopt a parameterization method based on Hamilton quaternion representation \cite{sola2017quaternion} and stereographic projection, which exhibits excellent numerical properties such as accelerated convergence as shown in \cite{terzakis2014quaternion}, to make trajectory optimization more efficient.
Utilizing the homeomorphism between the hyperplane $\mathbb{R}^3$ and the hypersphere $\mathbb{S}^3$ with one pole removed, a stereographic projection maps an arbitrary vector $\bm{\sigma} \in \mathbb{R}^3$ as a unit quaternion $\mathbf{Q} := \begin{bmatrix}q_w & q_x & q_y & q_z\end{bmatrix}^\top = \begin{bmatrix}q_w & \mathbf{r}^{\top}\end{bmatrix}^\top$ representing the rotation 
\begin{equation}
  \mathbf{R}(\mathbf{Q}) := 
  (q_w^2 - \mathbf{r}^\top \mathbf{r}) \mathbf{I} + 2 \mathbf{r} \mathbf{r}^\top + 2 q_w \mathbf{r}^\wedge 
\end{equation}

If the pole is chosen as $\mathbf{Q}_{N} = \begin{bmatrix}1 & 0 & 0 & 0\end{bmatrix}^\top$,
the map is expressed as follows:
\begin{equation}
    \mathbf{Q}(\bm{\sigma}) := 
    \begin{bmatrix}
        \frac{\bm{\sigma}^{\top}\bm{\sigma} - 1}{\bm{\sigma}^{\top}\bm{\sigma} + 1} &
        \frac{2\bm{\sigma}^{\top}}{\bm{\sigma}^{\top}\bm{\sigma} + 1}
    \end{bmatrix}^{\top} \in \mathbb{S}^3 \backslash \{\mathbf{Q}_N\}, \forall \bm{\sigma} \in \mathbb{R}^3.
    \label{equ:stereographic_projection}
\end{equation}

We can see that $\mathbf{Q}(\bm{\sigma}):\mathbb{R}^3 \rightarrow \mathbb{S}^3 \backslash \{\mathbf{Q}_N\}$ is smooth and one-to-one.
Thus each rotation has at most two distinct $\bm{\sigma}$ counterparts ($\mathbf{R} = \mathbf{I}$ corresponds only to the origin of $\mathbb{R}^3$), 
which greatly reduces the possibility of ambiguity.

The angular velocity expressed in $\mathcal{F}_W$ can be calculated as:
\begin{equation}
    \bm{\omega} = 2\mathbf{U}\dot{\mathbf{Q}} = 2\mathbf{U}\mathbf{G}^{\top}\dot{\bm{\sigma}},
    \label{equ:omega}
\end{equation}
where $\mathbf{U} := \begin{bmatrix} -\mathbf{r} & w\mathbf{I} + \mathbf{r}^{\wedge}\end{bmatrix} \in \mathbb{R}^{3 \times 4}$
 and $\mathbf{G} := 
 \begin{bmatrix}
     \frac{\partial q_w}{\partial \bm{\sigma}} & 
     \frac{\partial q_x}{\partial \bm{\sigma}} & 
     \frac{\partial q_y}{\partial \bm{\sigma}} & 
     \frac{\partial q_z}{\partial \bm{\sigma}}
 \end{bmatrix} \in \mathbb{R}^{3 \times 4}$.
Then, we can calculate the rotation-related penalty terms.

\begin{figure}[t]
    \centering
    \subfloat[]{\includegraphics[width=0.23\textwidth]{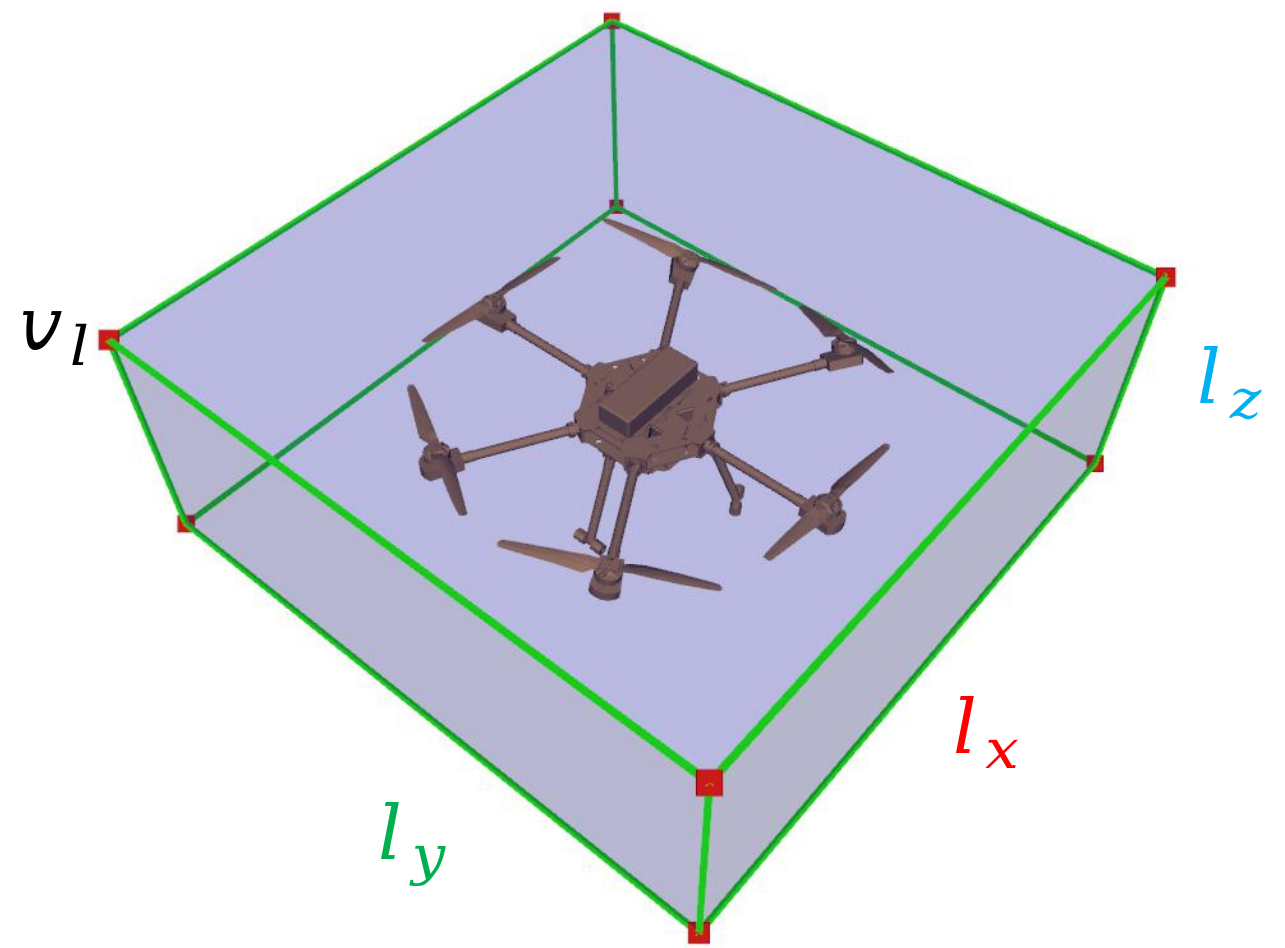}\label{subfig:approximate_cuboid}}
    \hfil
    \subfloat[]{\includegraphics[width=0.23\textwidth]{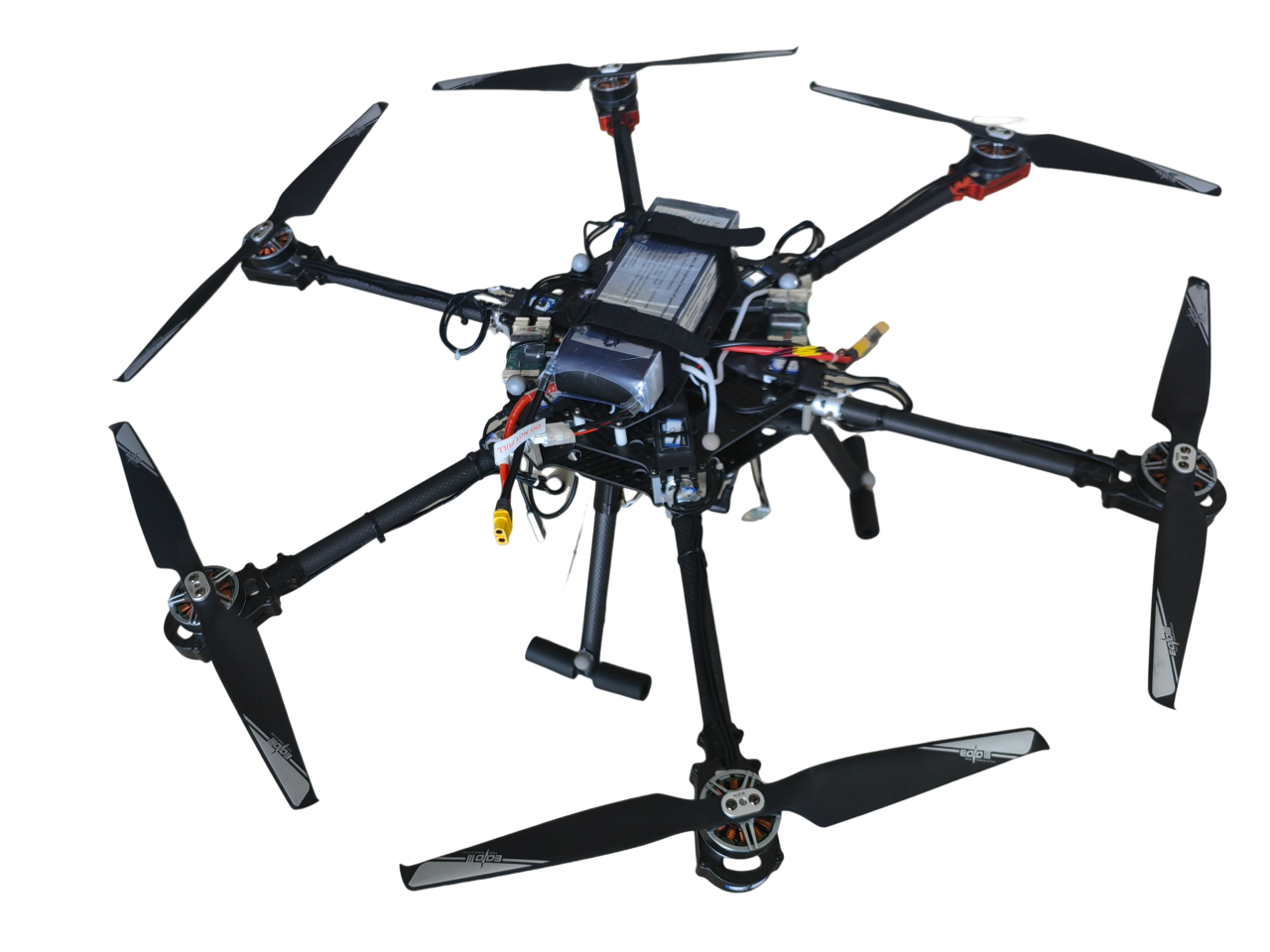}\label{subfig:real_omni_hex}}
    \caption{Illustrations of some experiment settings. (a) shows the cuboid used to approximate the vehicle's shape, of which the center coincides with the vehicle's CoM, and the three symmetry axes are parallel with axes of $\mathcal{F}_b$. (b) shows the real OmniHex we develop. }
    \label{fig:exp_platform}
\end{figure}

\begin{table}[t]
    \caption{Optimization Settings}
    \label{tab:optimization_settings}
    \centering
    \begin{tabular}{|c|c|c|c|}
    \hline
    $\kappa$ & $v_{\text{max}}$ & $a_{\text{max}}$ & $\omega_{\text{max}}$ \\
    \hline
    16 & $0.6 \text{m}\cdot\text{s}^{-1}$ & $2.0 \text{m}\cdot\text{s}^{-2}$ & $0.5 \text{rad}\cdot\text{s}^{-1}$ \\
    \hline
    \hline
    $\mathcal{W}_v$ & $\mathcal{W}_a$ & $\mathcal{W}_\omega$ & $k_\rho$ \\
    \hline
     $1\times10^5$ & $1\times10^5$ & $1\times10^5$ & 10 \\
    \hline
    \end{tabular}
\end{table}

\begin{figure}[ht]
    \centering
    \subfloat[]{\includegraphics[width=3.1in]{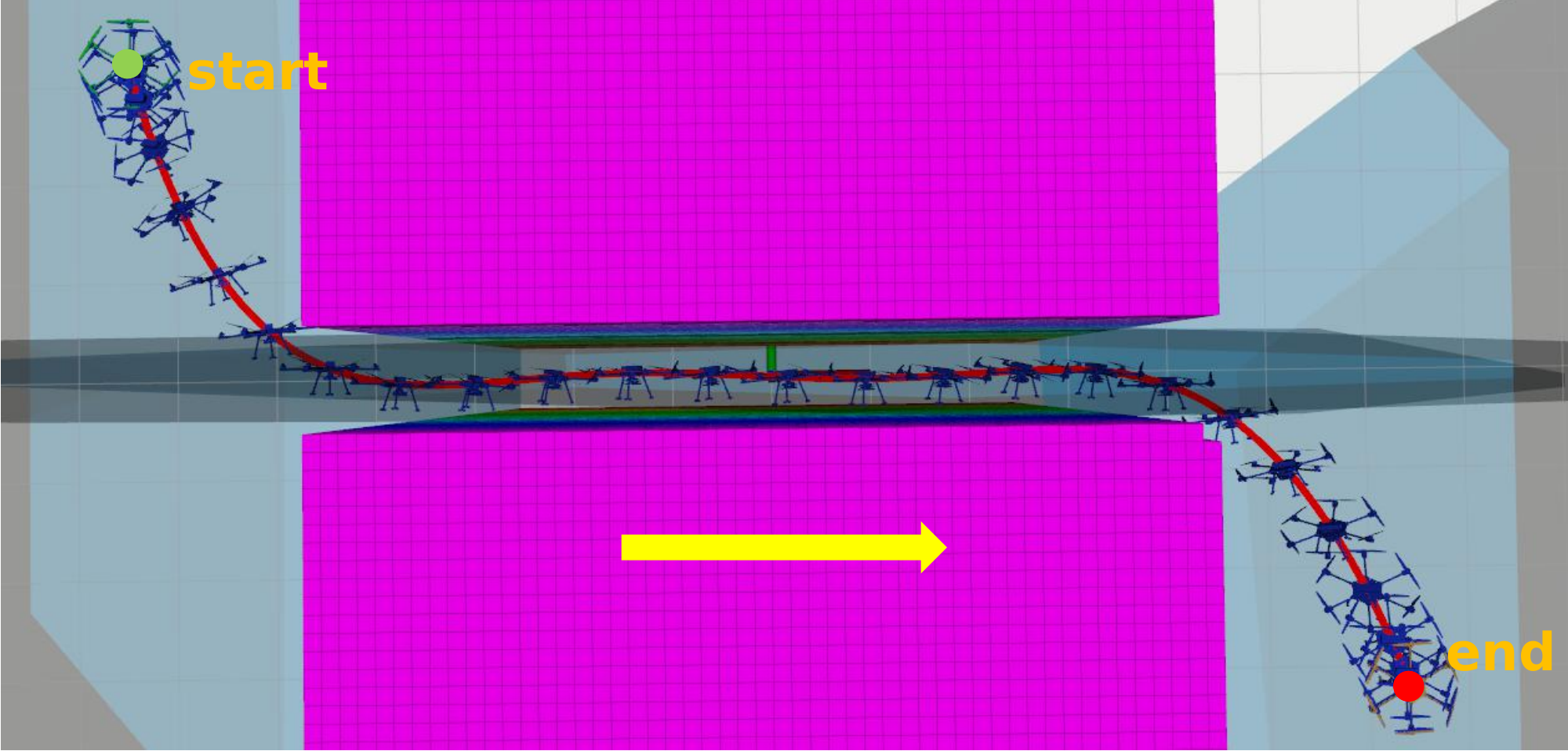}
    \label{subfig:traj_gen_narrow_passage}}
    \hfil
    \subfloat[]{\includegraphics[width=3.1in]{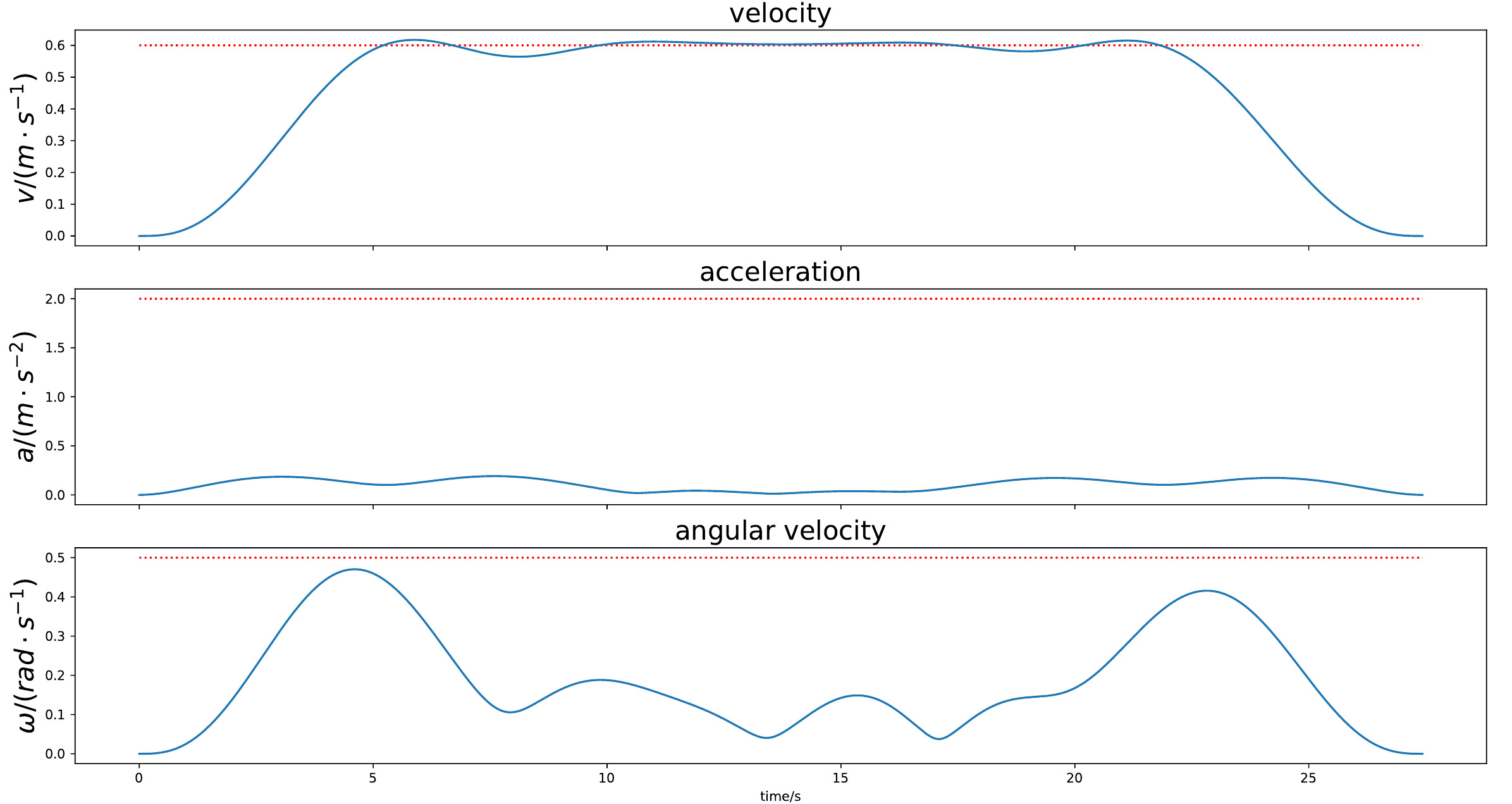}
    \label{subfig:dyn_prop_2D_narrow_passage}}
    
    \caption{A 6-D trajectory generated in Scenario A. (a) shows the overview of the map and the trajectory (from the top view), where a series of light blue transparent convex polyhedra form the SFC, 
    the red curve represents the geometry of the position part of the 6-D trajectory (i.e., the trajectory of CoM), 
    and the vehicle models in dark blue represent the rotation parts of the 6-D trajectory at the corresponding sampling points. (b) shows the kinodynamic properties of the trajectory, specifically the changes of $\Vert \mathbf{v} \Vert_2$, $\Vert \mathbf{a} \Vert_2$, and $\Vert \bm{\omega} \Vert_2$ w.r.t. time (shown by the blue curves), and the corresponding maximum limits (shown by the red dashed lines). }
    \label{fig:results_narrow_passage}
\end{figure}

\begin{figure}[ht]
    \centering
    \subfloat[]{\includegraphics[width=3.1in]{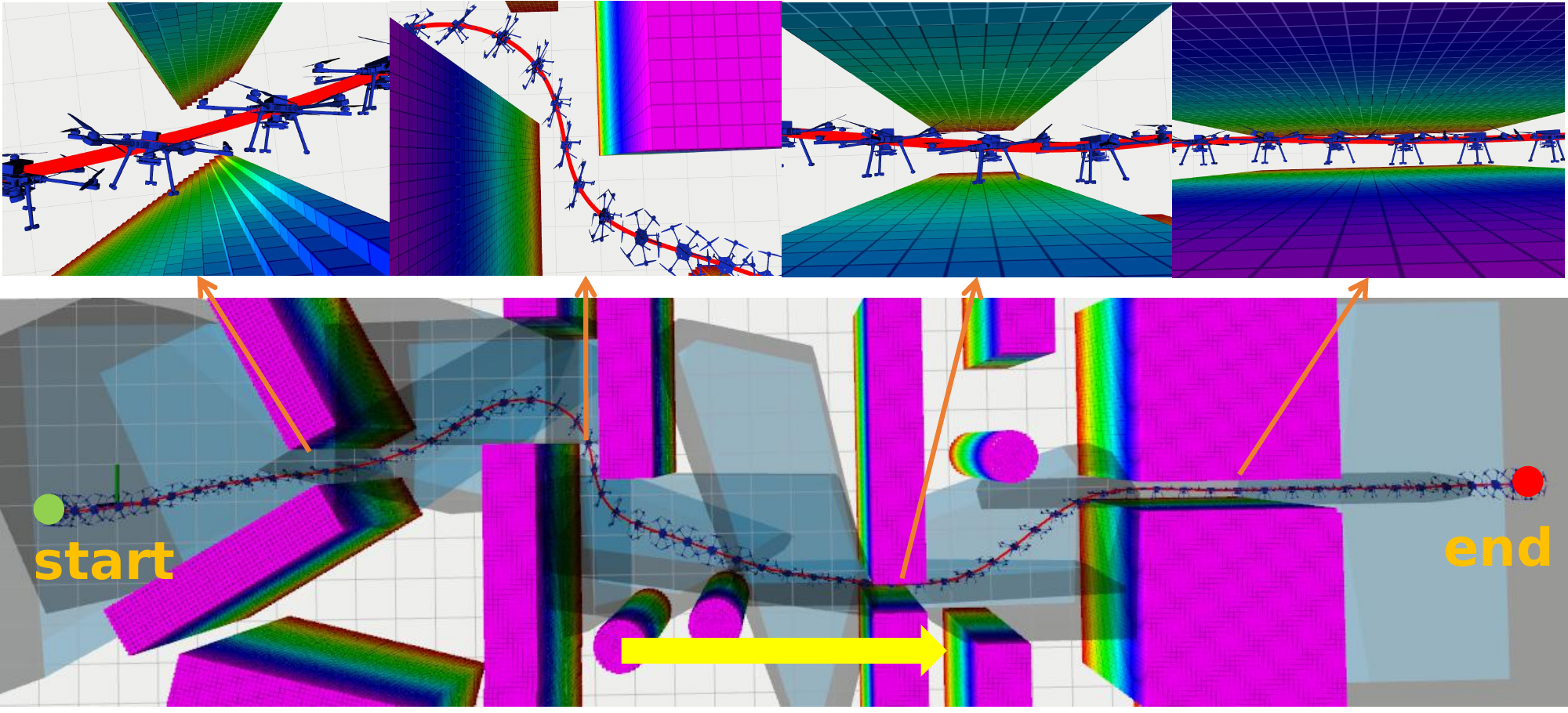}%
    \label{subfig:traj_gen_cluttered_env}}
    \hfil
    \subfloat[]{\includegraphics[width=3.1in]{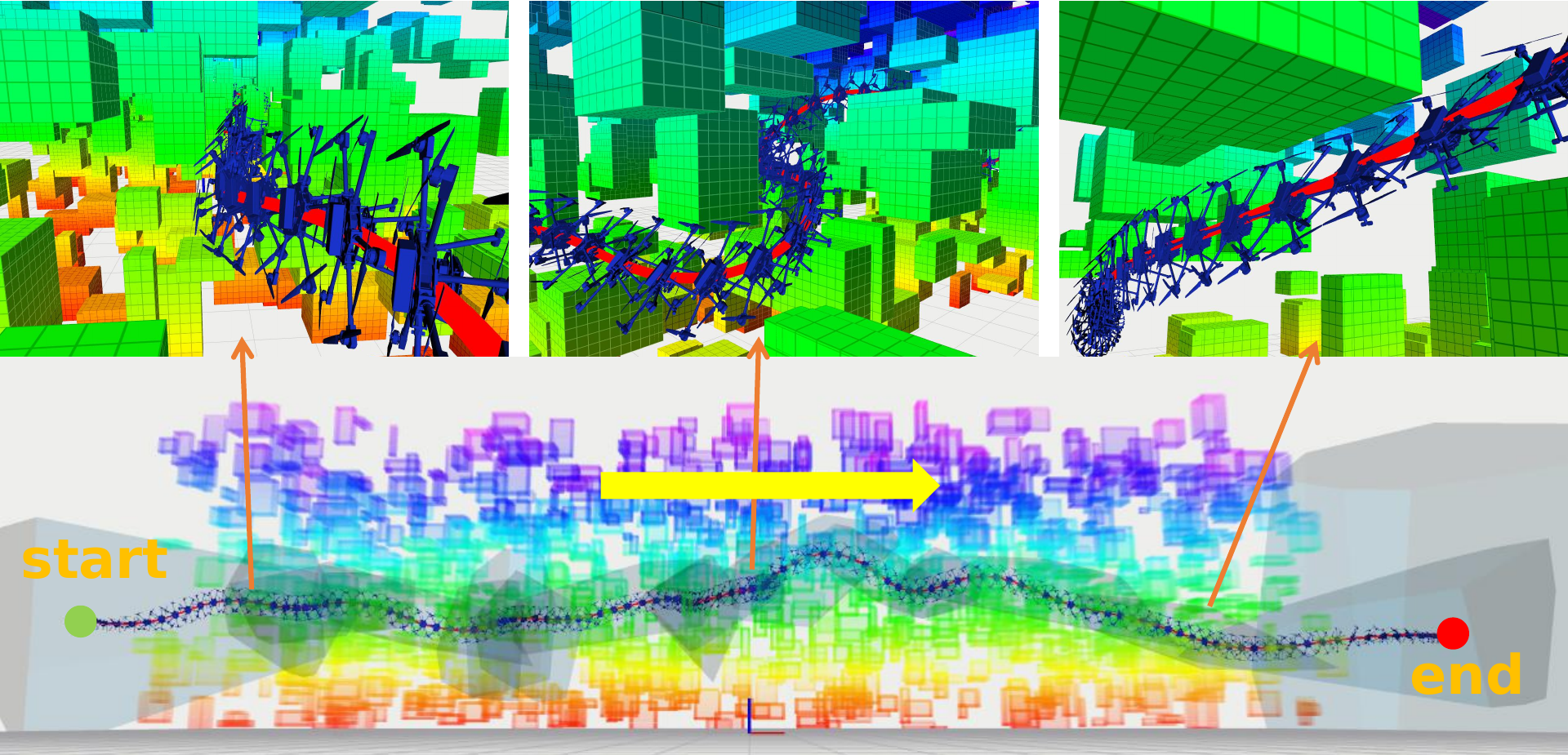}%
    \label{subfig:traj_gen_random_floating_map}}
    \hfil
    
    \iffalse
    \subfloat[]{\includegraphics[width=0.48\textwidth]{figures/dyn_prop_2D_cluttered_env.pdf}%
    \label{subfig:dyn_prop_2D_cluttered_env}}
    \hfil
    \subfloat[]{\includegraphics[width=0.48\textwidth]{figures/dyn_prop_2D_random_floating_map.pdf}%
    \label{subfig:dyn_prop_2D_random_floating_map}}
    \hfil
    \fi
    \caption{Trajectory generation results. (a) shows a trajectory generated in Scenario B (from the top view), and (b) shows that in Scenario C (from the side view). We adjusted the transparency of the point cloud visualization in the bottom image of (b) so that the generated SFC and 6-D trajectory can be displayed more clearly.}
    \label{fig:results_cluttered_env_and_random_floating_map}
\end{figure} 

\begin{figure}[ht]
    \centering
    \includegraphics[width=0.48\textwidth]{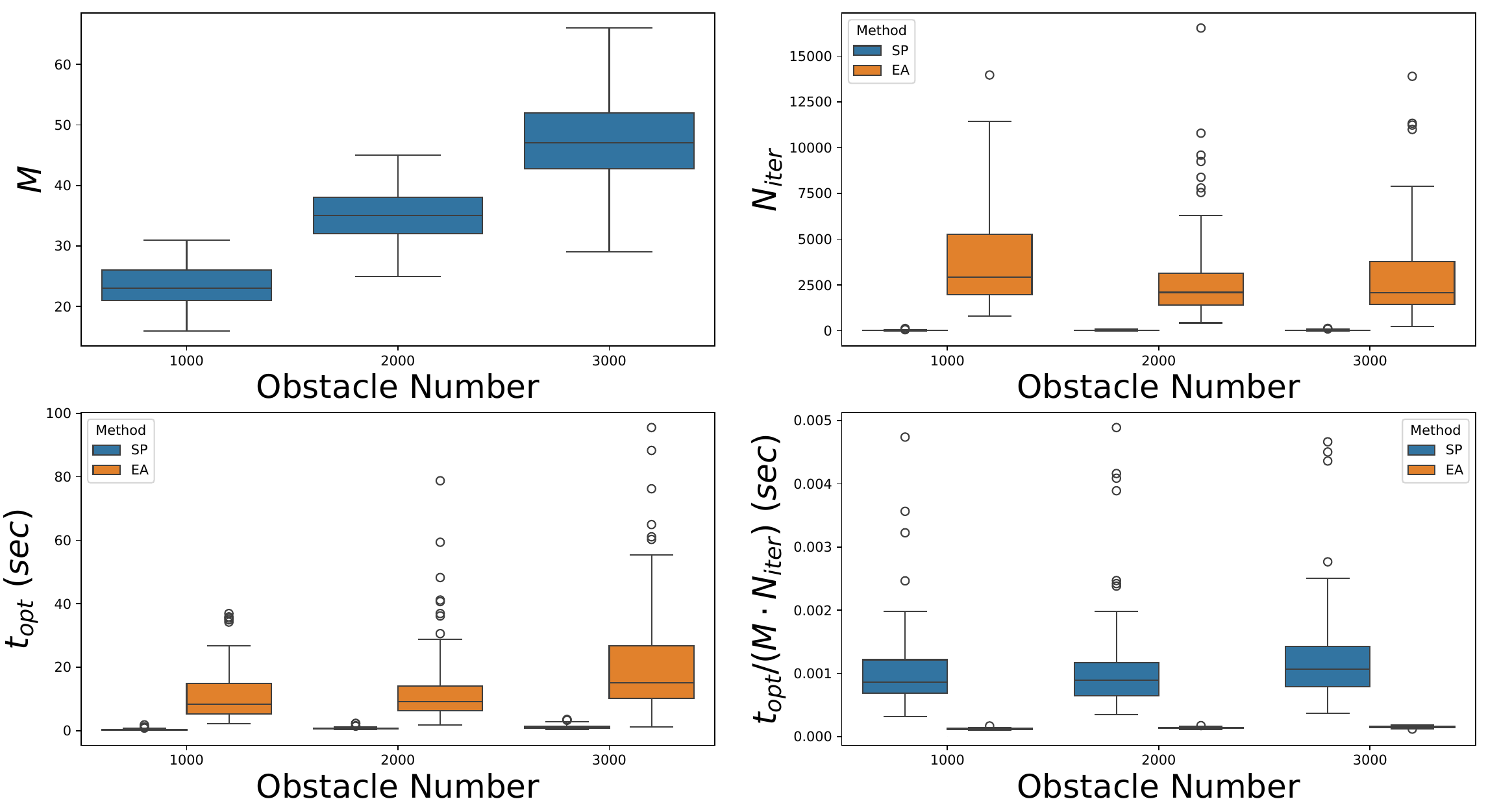}%
    \caption{The efficiency comparison between stereographic-projection-based (SP) method and Euler-angle-based (EA) method. The circles represent outliers outside the range $[Q_1-2 \cdot IQR, Q_3 + 2 \cdot IQR]$, where $Q_1$ is the lower quartile, $Q_3$ is the upper quartile, and $IQR = Q_3 - Q_1$. }
\label{fig:comparison_result}
\end{figure}

\begin{figure*}[ht]
    \centering
    \includegraphics[width=\textwidth]{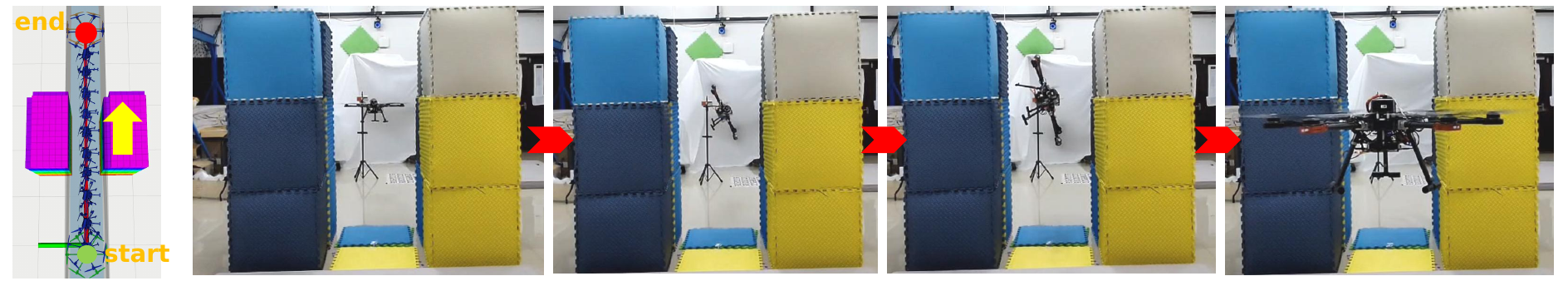}%
    \caption{OmniHex executes the generated trajectory to fly through a narrow passage. The leftmost figure is the visualization of the generated trajectory, and the others are snapshots of the experiment. The red arrow indicates the chronological order of the snapshots. }
\label{fig:real_traj_1}
\end{figure*}

\begin{figure}[ht]
    \centering
    \includegraphics[width=3.5in]{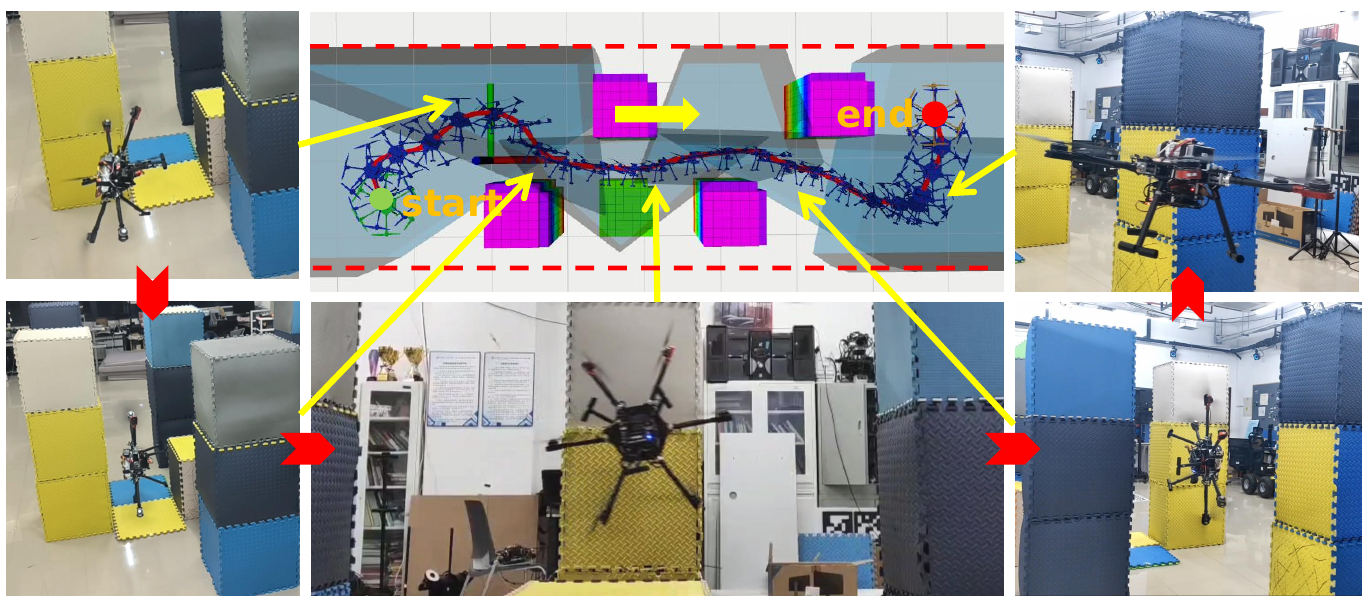}%
    \caption{OmniHex executes the generated trajectory to fly through a narrow environment. The top middle figure shows the generated trajectory where the red dashed lines represent the map's boundaries. }
\label{fig:real_traj_2}
\end{figure}

\section{Simulations and Experiments}
In this section, we show the performance of the proposed method. 
First, we generate trajectories in cluttered simulation environments to verify the effectiveness and efficiency of our method. 
Then, real-world experiments are carried out to test the practicability of our method. 
%This section takes a tilt-rotor omnidirectional hexarotor vehicle (hereinafter called OmniHex) as the research object. 
%Firstly, the trajectory generation experiment is carried out in virtual environments with randomly distributed obstacles to test our method's obstacle avoidance performance and computational efficiency.
%Then we generate a 6-D trajectory in a complex simulation environment through several narrow passages in succession 
%and let the simulation model track it to verify the practicality of our method.

\subsection{Implementation Details}
The initial path is obtained by applying RRT to the $SE(3)$ configuration space and is used to generate SFC using RILS.
The algorithm used to solve the optimization problem \eqref{equ:final_uncons_opt_problem} is L-BFGS \cite{liu1989limited}, 
with the backtracking method \cite{boyd2004convex} used for line search. 
We implement all the trajectory generation algorithms in C++17 using a single thread. 
The hardware platform used for simulations is a Dell G5 laptop with an Intel Core i7-10750H CPU @ 2.60GHz running the Ubuntu 20.04 operating system.
The trajectory generation algorithms are all executed sequentially without explicit hardware acceleration. 

\subsection{Parameter Settings} \label{subsec:parameter_settings}

To achieve whole-body obstacle avoidance, as Section \ref{subsec:safety_constraint} mentions, we approximate the vehicle's shape as a convex polyhedron that envelops the body.
Here we use a cuboid (Fig. \ref{subfig:approximate_cuboid}) whose dimensions are $l_x = l_y = 1.1\text{m},l_z = 0.42\text{m}$, according to the size of OmniHex. 
The coordinates of the eight vertices in $\mathcal{F}_b$ are $\tilde{\bm{v}}_l = \begin{bmatrix}\pm \frac{l_x}{2} & \pm \frac{l_y}{2} & \pm \frac{l_z}{2}\end{bmatrix}^\top$. 
Note that it is also possible to utilize a shape with closer proximity, such as a hexagon. 
However, the extent to which the solution space can be expanded using this type of shape rather than a cuboid is very limited, and more vertices may require additional computation.
We set $s = 4$ to ensure the smoothness of control inputs according to \eqref{equ:Psi_u}. 
The initial value of $\mathbf{q}^{\sigma}$ of trajectory optimization is set to $\mathbf{0}$. 
The other optimization settings for simulations are listed in Table \ref{tab:optimization_settings}.

\subsection{Simulation Results}
In this section, simulations are carried out in three virtual scenarios to test the proposed method's effectiveness and efficiency. 
We generate a point cloud map for each scenario to provide obstacle information for our trajectory generation framework. 
The three virtual scenarios are described as follows. 

\textbf{Scenario A: }We expect the vehicle to fly through a narrow, straight, vertical passage. The passage is 0.7m wide, which is narrower than the vehicle's width ($l_x$ and $l_y$) and wider than its thickness ($l_z$). 

\textbf{Scenario B: }An unstructured cluttered environment with extremely narrow regions, including the narrow passage described in Scenario A. The map is restricted to a $40\text{m}\times10\text{m}\times6\text{m}$ area. 

\textbf{Scenario C: }We expect the vehicle to pass through a $30\text{m}\times10\text{m}\times8\text{m}$ area containing 1000 randomly distributed floating cuboid obstacles. 

Since there is no available 6-D trajectory generation baseline  considering obstacles for OMAVs, 
we solely present the performance of our method.

\begin{table}[t]
    \caption{Efficiency Evaluation. }
    \label{tab:efficiency_evaluation}
    \centering
    \resizebox{\columnwidth}{!}{
        \begin{tabular}{cccccccc}
            \hline
            scenario & $M_\mathcal{P}$ & $M$ & $D$ &  $t_\text{rrt}$ & $t_\text{sfc}$ & $N_\text{iter}$ & $t_\text{opt}$ \\
            \hline
            A & 4 & 7 & 11.2m & 598ms & 4ms & 60 & 88ms \\ 
            % \hline
            B & 12 & 20 & 34.0m & 1284ms & 64ms & 81 & 611ms \\ 
            % \hline 
            C & 19 & 24 & 38.1m & 55ms & 120ms & 48 & 422ms \\
            \hline
            \multicolumn{8}{p{\columnwidth}}{\textbf{Note: }
            % $M_\mathcal{P}$ is the number of polyhedron primitives in the SFC. $M$ is the number of trajectory pieces. $D$ is the straight-line distance between a trajectory's start and end positions. 
            $t_\text{rrt}$ is the time it takes to find an initial feasible path using RRT. $ t_\text{sfc}$ is the time it takes to generate an SFC using RILS based on the initial path. $t_\text{opt}$ is the time it takes to solve the trajectory optimization problem, and $N_\text{iter}$ is the iteration number of L-BFGS. }
        \end{tabular}
    }
\end{table}

Fig. \ref{subfig:traj_gen_narrow_passage} shows the 6-D trajectory generated in Scenario A and the corresponding kinodynamic properties under the above settings. 
The generated trajectory allows the vehicle to smoothly and appropriately change its attitude to adapt to the narrow space in the passage, avoiding collision while flying toward the target. 
Fig. \ref{subfig:dyn_prop_2D_narrow_passage} presents the constrained kinodynamic properties.
Although the constraints are relaxed, they are effectively satisfied in the resulting trajectory. 
Moreover, the velocity norm reaches $v_{\text{max}}$ most of the time, which shows that our method can fully exploit the performance of OMAVs. 
Note that due to the softening step, the resulting trajectory may slightly exceed the constraints \cite{yang2021whole}.
Nonetheless, in most cases the given constraints can be well satisfied and we can handle the possible violation by carefully tuning the optimizer parameters or by adding a certain margin when approximating the shape of the robot in practice. 

Fig. \ref{fig:results_cluttered_env_and_random_floating_map} shows the trajectories generated by the proposed method in scenarios B and C, which are much more challenging than Scenario A. 
The vehicle has been successfully constrained in the SFC throughout the journey and can change its attitude flexibly to avoid obstacles. 
The simulation results show that our method allows OMAVs to fly safely even in extremely cluttered environments. 
Due to space limitation, we are unableto give the corresponding kinodynamic properties of Fig. \ref{fig:results_cluttered_env_and_random_floating_map} in this manuscript.
Please refer to our video attachment for more details. 

For efficiency evaluation, we list the computation times of the results in Fig. \ref{fig:results_narrow_passage} and \ref{fig:results_cluttered_env_and_random_floating_map} in Table \ref{tab:efficiency_evaluation}. 
$t_\text{rrt}$ has a relatively large randomness, and in general, the narrower the feasible space, the larger $t_\text{rrt}$ tends to be. 
$ t_\text{sfc}$ is positively correlated with the number of path segments and the point cloud size. 
Since there are alternative methods available for the initial path search and the SFC generation that can be plug and play, we do not discuss the impact of these two stages on the efficiency of our framework here. 
The time complexity of each iteration (where the value of the objective function and the related gradients are calculated) is $O(M)$ \cite{wang2022geometrically}, so $t_\text{opt}$ is approximately proportional to $M$ and $N_\text{iter}$. 
The trajectory optimization stage shows relatively high solving efficiency, comparable to the performance on the CPU in \cite{han2021fast}, while \cite{han2021fast} generates only 3-D position trajectories for underactuated MAVs.
Therefore, our framework have the application potential in scenarios requiring real-time performance. 

In order to demonstrate the efficiency of stereographic-projection-based rotation parameterization (SP method), we also implement the Euler-Angle-based method (EA method) for additional comparison. 
The comparison experiment is carried out in a virtual environment that is basically the same as Scenario C, except that we set three cases for the number of obstacles: 1000, 2000, and 3000. For each case, we repeat the following steps 100 times: generate a point cloud map and an SFC, and then utilize our trajectory optimizer to perform trajectory optimization within that SFC using both SP and EA methods under the same optimization settings. 
Therefore, the number of trajectory pieces obtained by the two methods is equal in each trial.
The results are presented in Figure \ref{fig:comparison_result}. 
From $t_\text{opt}$ we can see that the SP method exhibits significantly higher efficiency than the EA method with a more concentrated time distribution and much fewer iterations for convergence. %Moreover, the EA method shows a much lower $t_\text{opt} / (M \cdot N_\text{iter})$ value, indicating that SP method requires fewer iterations for convergence. 
In addition, an increase in the obstacle number will increase obstacle distribution density, leading to an increasing detour of the trajectories, which is consistent with the general increase in $M$ with obstacle number as shown in Figure \ref{fig:comparison_result}. 
On the other hand, $t_\text{opt} / (M \cdot N_\text{iter})$ is generally stable, which is consistent with the linear complexity of the trajectory optimization algorithm. 
In general, compared with EA method, SP method has higher convergence speed and is more suitable for real-time applications.

\subsection{Real-World Experiments}

The results presented in this section aim to verify the applicability of the proposed method can be applied to real OMAVs. 
We set up cuboid obstacles with known sizes and positions in the environment. Then, we generate collision-free 6-D trajectories in advance using the proposed method. 
The experimental platform is the OmniHex (Fig. \ref{subfig:real_omni_hex}) mentioned in Section \ref{subsec:control}. 
Six Dynamixel XH430-W210-T servos each provide one rotor's degree of tilt freedom. 
%A PixHawk Cube Orange running the modified PX4 Autopilot \cite{meier2015px4} is used as the flight controller. 
%An onboard Intel NUC11PAHi7 computer sends 6-D trajectory setpoints to the controller via a serial port. 
%A 1000mAh/14.8V LiPo battery is used to power the entire vehicle. 
An OptiTrack motion capture system provides high-precision real-time 6-D pose feedback to the vehicle at 100Hz via WiFi. 
For the trajectory optimization settings, we set the kinodynamic constraints according to the real vehicle's performance as : $v_{\max} = 0.36 \text{m}\cdot\text{s}^{-1}, a_{\max} = 0.5 \text{m}\cdot\text{s}^{-2}, \omega_{\max} = 1.0 \text{rad}\cdot\text{s}^{-1}$. 

Fig. \ref{fig:real_traj_1} shows OmniHex executing the generated trajectory, flying through a narrow passage that is 0.7m wide and 1.2m long. OmniHex must tilt a large angle to fly through the narrow passage without collision. 
Fig. \ref{fig:real_traj_2} shows OmniHex executing the generated trajectory, flying through an area with several cuboid obstacles. 
OmniHex accurately follows the reference trajectories and flies smoothly and safely from the starting points to the target points. 
The real-world experiment results demonstrate the practicability of our method. 
Please refer to our video attachment for more details.

\section{Conclusion}
We present a computationally efficient 6-D trajectory generation framework which can fully exploit the obstacle avoidance potential of OMAVs. 
A 6-D trajectory optimization problem considering safety and kinodynamic constraints is formulated.
A rational quaternion-based rotation parameterization method is adopted to achieve efficient optimization and high-quality solution. 
Simulations and real-world experiments are carried out to verify the performance of our method. 
Furthermore, our method can be applied to any platform that can do free and controlled 6-D rigid body motion, including spacecraft. 

The primary limitation of our method is its lack of consideration of possible dynamic changes of the environment, such as pedestrians, which limits its scope of application. In addition, utilizing the independence of each sampling point in the penalty function, the computational efficiency of our method can be further improved with parallel computation.
Moreover, we hope to enable autonomous OMAV-based aerial manipulation based on the proposed method, as well as the onboard sensing system. 
Achieving the above improvements and goals will be our future work.

\bibliographystyle{IEEEtran}
\bibliography{root}

\begin{thebibliography}{10}
\providecommand{\url}[1]{#1}
\csname url@rmstyle\endcsname
\providecommand{\newblock}{\relax}
\providecommand{\bibinfo}[2]{#2}
\providecommand\BIBentrySTDinterwordspacing{\spaceskip=0pt\relax}
\providecommand\BIBentryALTinterwordstretchfactor{4}
\providecommand\BIBentryALTinterwordspacing{\spaceskip=\fontdimen2\font plus
\BIBentryALTinterwordstretchfactor\fontdimen3\font minus
  \fontdimen4\font\relax}
\providecommand\BIBforeignlanguage[2]{{%
\expandafter\ifx\csname l@#1\endcsname\relax
\typeout{** WARNING: IEEEtran.bst: No hyphenation pattern has been}%
\typeout{** loaded for the language `#1'. Using the pattern for}%
\typeout{** the default language instead.}%
\else
\language=\csname l@#1\endcsname
\fi
#2}}

\bibitem{brescianini2016design}
D.~Brescianini and R.~D'Andrea, ``Design, modeling and control of an
  omni-directional aerial vehicle,'' in \emph{2016 IEEE international
  conference on robotics and automation (ICRA)}.\hskip 1em plus 0.5em minus
  0.4em\relax IEEE, 2016, pp. 3261--3266.

\bibitem{kamel2018voliro}
M.~Kamel, S.~Verling, O.~Elkhatib, C.~Sprecher, P.~Wulkop, Z.~Taylor,
  R.~Siegwart, and I.~Gilitschenski, ``The voliro omniorientational hexacopter:
  An agile and maneuverable tiltable-rotor aerial vehicle,'' \emph{IEEE
  Robotics \& Automation Magazine}, vol.~25, no.~4, pp. 34--44, 2018.

\bibitem{morbidi2018energy}
F.~Morbidi, D.~Bicego, M.~Ryll, and A.~Franchi, ``Energy-efficient trajectory
  generation for a hexarotor with dual-tilting propellers,'' in \emph{2018
  IEEE/RSJ International Conference on Intelligent Robots and Systems
  (IROS)}.\hskip 1em plus 0.5em minus 0.4em\relax IEEE, 2018, pp. 6226--6232.

\bibitem{pantic2021mesh}
M.~Pantic, L.~Ott, C.~Cadena, R.~Siegwart, and J.~Nieto, ``Mesh manifold based
  riemannian motion planning for omnidirectional micro aerial vehicles,''
  \emph{IEEE Robotics and Automation Letters}, vol.~6, no.~3, pp. 4790--4797,
  2021.

\bibitem{mellinger2011minimum}
D.~Mellinger and V.~Kumar, ``Minimum snap trajectory generation and control for
  quadrotors,'' in \emph{2011 IEEE international conference on robotics and
  automation}.\hskip 1em plus 0.5em minus 0.4em\relax IEEE, 2011, pp.
  2520--2525.

\bibitem{gao2017gradient}
F.~Gao, Y.~Lin, and S.~Shen, ``Gradient-based online safe trajectory generation
  for quadrotor flight in complex environments,'' in \emph{2017 IEEE/RSJ
  international conference on intelligent robots and systems (IROS)}.\hskip 1em
  plus 0.5em minus 0.4em\relax IEEE, 2017, pp. 3681--3688.

\bibitem{zhou2020ego}
X.~Zhou, Z.~Wang, H.~Ye, C.~Xu, and F.~Gao, ``Ego-planner: An esdf-free
  gradient-based local planner for quadrotors,'' \emph{IEEE Robotics and
  Automation Letters}, vol.~6, no.~2, pp. 478--485, 2020.

\bibitem{gao2020teach}
F.~Gao, L.~Wang, B.~Zhou, X.~Zhou, J.~Pan, and S.~Shen, ``Teach-repeat-replan:
  A complete and robust system for aggressive flight in complex environments,''
  \emph{IEEE Transactions on Robotics}, vol.~36, no.~5, pp. 1526--1545, 2020.

\bibitem{gao2019flying}
F.~Gao, W.~Wu, W.~Gao, and S.~Shen, ``Flying on point clouds: Online trajectory
  generation and autonomous navigation for quadrotors in cluttered
  environments,'' \emph{Journal of Field Robotics}, vol.~36, no.~4, pp.
  710--733, 2019.

\bibitem{yang2021whole}
S.~Yang, B.~He, Z.~Wang, C.~Xu, and F.~Gao, ``Whole-body real-time motion
  planning for multicopters,'' in \emph{2021 IEEE International Conference on
  Robotics and Automation (ICRA)}.\hskip 1em plus 0.5em minus 0.4em\relax IEEE,
  2021, pp. 9197--9203.

\bibitem{han2021fast}
Z.~Han, Z.~Wang, N.~Pan, Y.~Lin, C.~Xu, and F.~Gao, ``Fast-racing: An
  open-source strong baseline for $se(3)$ planning in autonomous drone
  racing,'' \emph{IEEE Robotics and Automation Letters}, vol.~6, no.~4, pp.
  8631--8638, 2021.

\bibitem{wang2022geometrically}
Z.~Wang, X.~Zhou, C.~Xu, and F.~Gao, ``Geometrically constrained trajectory
  optimization for multicopters,'' \emph{IEEE Transactions on Robotics},
  vol.~38, no.~5, pp. 3259--3278, 2022.

\bibitem{ren2023online}
Y.~Ren, S.~Liang, F.~Zhu, G.~Lu, and F.~Zhang, ``Online whole-body motion
  planning for quadrotor using multi-resolution search,'' in \emph{2023 IEEE
  International Conference on Robotics and Automation (ICRA)}.\hskip 1em plus
  0.5em minus 0.4em\relax IEEE, 2023, pp. 1594--1600.

\bibitem{lavalle1998rapidly}
S.~LaValle, ``Rapidly-exploring random trees: A new tool for path planning,''
  \emph{Research Report 9811}, 1998.

\bibitem{sola2017quaternion}
J.~Sola, ``Quaternion kinematics for the error-state kalman filter,''
  \emph{arXiv preprint arXiv:1711.02508}, 2017.

\bibitem{fliess1995flatness}
M.~Fliess, J.~L{\'e}vine, P.~Martin, and P.~Rouchon, ``Flatness and defect of
  non-linear systems: introductory theory and examples,'' \emph{International
  journal of control}, vol.~61, no.~6, pp. 1327--1361, 1995.

\bibitem{bodie2020towards}
K.~Bodie, Z.~Taylor, M.~Kamel, and R.~Siegwart, ``Towards efficient full pose
  omnidirectionality with overactuated mavs,'' in \emph{Proceedings of the 2018
  International Symposium on Experimental Robotics}.\hskip 1em plus 0.5em minus
  0.4em\relax Springer, 2020, pp. 85--95.

\bibitem{liu2017planning}
S.~Liu, M.~Watterson, K.~Mohta, K.~Sun, S.~Bhattacharya, C.~J. Taylor, and
  V.~Kumar, ``Planning dynamically feasible trajectories for quadrotors using
  safe flight corridors in 3-d complex environments,'' \emph{IEEE Robotics and
  Automation Letters}, vol.~2, no.~3, pp. 1688--1695, 2017.

\bibitem{terzakis2014quaternion}
G.~Terzakis, P.~Culverhouse, G.~Bugmann, \emph{et~al.}, ``On quaternion based
  parametrization of orientation in computer vision and robotics,'' 2014.

\bibitem{liu1989limited}
D.~C. Liu and J.~Nocedal, ``On the limited memory bfgs method for large scale
  optimization,'' \emph{Mathematical programming}, vol.~45, no. 1-3, pp.
  503--528, 1989.

\bibitem{boyd2004convex}
S.~P. Boyd and L.~Vandenberghe, \emph{Convex optimization}.\hskip 1em plus
  0.5em minus 0.4em\relax Cambridge university press, 2004.

\end{thebibliography}

%%%%%%%%%%%%%%%%%%%%%%%%%%%%%%%%%%%%%%%%%%%%%%%%%%%%%%%%%%%%%%%%%%%%%%%%%%%%%%%%

\end{document}